# Standalone and RTK GNSS on 30,000 km of North American Highways

Tyler G. R. Reid*, Nahid Pervez*, Umair Ibrahim**, Sarah E. Houts**, Gaurav Pandey*, Naveen K. R. Alla*, Andy Hsia*

*Ford Motor Company
**Ford Autonomous Vehicles, LLC

**ABSTRACT**

There is a growing need for vehicle positioning information to support Advanced Driver Assistance Systems (ADAS), Connectivity (V2X), and Automated Driving (AD) features. These range from a need for road determination (<5 meters), lane determination (<1.5 meters), and determining where the vehicle is within the lane (<0.3 meters). This work examines the performance of Global Navigation Satellite Systems (GNSS) on 30,000 km of North American highways to better understand the automotive positioning needs it meets today and what might be possible in the near future with wide area GNSS correction services and multi-frequency receivers. This includes data from a representative automotive production GNSS used primarily for turn-by-turn navigation as well as an Inertial Navigation System which couples two survey grade GNSS receivers with a tactical grade Inertial Measurement Unit (IMU) to act as ground truth. The latter utilized networked Real-Time Kinematic (RTK) GNSS corrections delivered over a cellular modem in real-time. We assess on-road GNSS accuracy, availability, and continuity. Availability and continuity are broken down in terms of satellite visibility, satellite geometry, position type (RTK fixed, RTK float, or standard positioning), and RTK correction latency over the network. Results show that current automotive solutions are best suited to meet road determination requirements at 98% availability but are less suitable for lane determination at 57%. Multi-frequency receivers with RTK corrections were found more capable with road determination at 99.5%, lane determination at 98%, and highway-level lane departure protection at 91%.

## INTRODUCTION

Vehicle location information is of increasing importance for Advanced Driver Assistance Systems (ADAS), connectivity (V2X), and Autonomous Driving (AD) features. Some features, such as turn-by-turn navigation instructions, provide convenience, where others, such as lane departure warning and lane-keeping, are safety-critical. Today, Global Navigation Satellite Systems (GNSS) primarily provide position information as a driver navigational aid [1]. However, new applications are driving more stringent localization requirements. These features are commonly broken down into categories requiring position information at the level of (i) which road, (ii) which lane, or (iii) where in the lane. Strict requirements on these categories have yet to be defined; however, Table 1 provides some data points. These range from <5 meters for road determination, <1.5 meters for lane determination, and <0.3 – 0.5 meters for lane departure warning and active control.

Road standards in the United States call for highway lanes to be 3.6 meters wide and local/city lanes to be at least 2.7 meters wide [2]. Hence, 'which road' positioning is generally taken as slightly better than two lane widths as this is the typical width of most two-way roads. Some applications that require this level of positioning are turn-by-turn navigation and geofencing. For example, Cadillac's Super Cruise level 2 self-driving feature is currently geofenced to limited access divided highways. Super Cruise utilizes precision Light Detection and Ranging (LiDAR) maps of select US and Canadian divided highways from Ushr [3] in conjunction with GNSS correction services from Trimble's RTX [4], [5] to achieve high confidence geofencing of the feature. This and other SAE level 2 self-driving features represent partial automation, where the human driver is responsible for monitoring the scene and the system is responsible for some dynamic driving tasks including steering, propulsion, and braking. The human driver must be ready to take over dynamic driving tasks immediately when the driver determines the system is incapable.

**Table 1: Summary of on-road positioning requirements.**

| Positioning Level | Accuracy [m] | Applications | Statistics | References |
|---|---|---|---|---|
| Which Road | < 5 | turn-by-turn navigation<br>geofencing | Undefined | [6], [7] |
| Which Lane | < 1.5 | V2X<br>ADAS | 1σ<br>Undefined | [6]–[8] |
| Where in Lane | < 0.5 highways<br>< 0.3 city roads | lane departure warning<br>autonomous driving | Undefined<br>$10^{-8}$ / h | [9], [10] |

The next step is 'which lane' positioning. Since most reasonably trafficked US roads are built with 3+ meter lane widths [2], better than half of this number, or 1.5 meters, is generally accepted as lane determination. The National Highway Safety Administration (NHTSA) has, as part of its Federal Motor Vehicle Safety Standards in Vehicle to Vehicle (V2V) Communications, determined that position must be reported to an accuracy of 1.5 meters (1σ or 68%) as this is tentatively believed to provide lane-level information for safety applications [8].

The most stringent positioning is that needed for maintaining the vehicle within its lane. This requirement is a combination of road geometry (width and curvature) along with vehicle dimensions. The small spaces between the vehicle edges and the painted lines on the road are the required position protection level to maintain the vehicle within its lane. For passenger vehicles in the US, highway road geometry requires <0.5 meters lateral positioning where local/city road geometry require <0.3 meters [9], [10]. For fully autonomous operation, these protection levels must be maintained to an integrity risk of $10^{-8}$ failures / hour of operation [9], [10]. This is equivalent to 99.999999% certainty in position.

Many localization technologies have been proposed to meet these needs. Relative navigation sensors based on LiDAR, computer vision, and others work by localizing to an a-priori map. This functions well in feature rich urban environments but can degrade in sparse highway settings. Qualitatively, positioning based on Global Navigation Satellite Systems is complementary, where a lack of features is typically synonymous with open skies and favorable satellite visibility. This paper focuses on the state of GNSS on predominantly highways in North America in 2018 and attempts to answer where it fits within ADAS, V2X, and Automated Driving (AD) both today and in the near future.

Between 2006 and 2009, the Vehicle Safety and Communications Applications (VSC-A) Project examined GNSS service availability as part of its scope in testing communications-based Vehicle-to-Vehicle (V2V) safety systems [11], [12]. The goal was to determine if Dedicated Short Range Communications (DSRC), in conjunction with vehicle positioning information, can improve vehicle safety systems. The final report was made available by NHTSA where the GNSS study was completed by the Position, Location, and Navigation (PLAN) group at the University of Calgary. This presented a GNSS availability literature review along with the results of an extensive testing campaign where 52 hours of on-road data was collected. This study was designed to investigate the accuracy and availability of positioning information from a VSC-like system under various conditions and configurations. This included various methods of relative positioning using V2V, Vehicle to Infrastructure (V2I), position differencing, and RTK in urban, suburban, and highway environments.

In 2010, on-road GPS availability was examined by Pilutti and Wallis [13] on 13,000 km of representative US roads comprising a real-world driving profile of freeways, rural, urban, and residential streets. Availability was characterized from the perspective of dilution of precision (DOP) and number of satellites in view where DOP outage duration, DOP distribution, and the distribution of the number of satellites available for positioning was reported.

Since 2010, the number of navigation satellites available has more than tripled, moving from GPS-only (USA) to the inclusion of a rebooted GLONASS (Russia) constellation in 2011 and now nearly completed Galileo (Europe) and BeiDou (China) constellations. This growth is shown in Figure 1 where there are now more than 100 operational navigation satellites in orbit. For the user, this implies a jump from 10 to 30+ satellites in view and has significant implications for on-road GNSS availability.

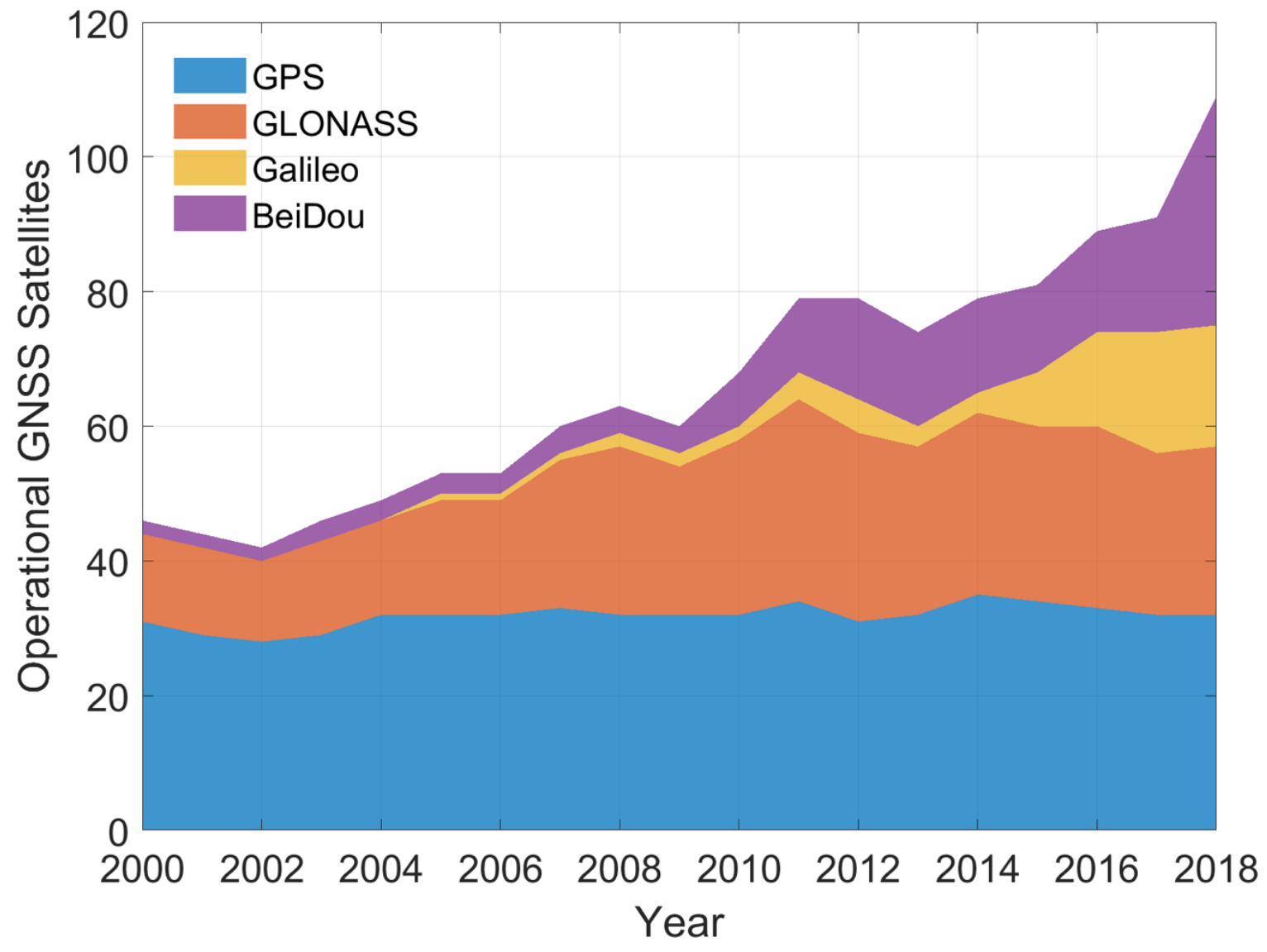

**Figure 1: Number of GNSS satellites as a function of time. The US GPS became available without selective availability for civilian use in 2000. The Russian GLONASS was again at full operational capability in 2011. The European Galileo and Chinese BeiDou are both nearing completion.**

Multi-constellation is just one of the advancements coming to the automotive domain, the others are multi-frequency and widespread corrections services. The use of these technologies is not new to the automotive domain. Survey grade GNSS receivers coupled with a tactical grade Inertial Measurement Unit (IMU) and correction services are often used in testing, validation, and research. One example is Stanford's Stanley, the winner of the 2006 Defense Advanced Research Projects Agency (DARPA) Grand Challenge, which made use of an OmniSTAR high precision correction-enabled GPS + IMU system for absolute positioning [14]. Traditionally such systems came at a high cost, however, we are now in a state of transition from specialized to volume markets and many technologies are rapidly maturing in large scale efforts to meet the demand for decimeter location in production. On the GNSS side, multi-frequency, mass market receivers are already here [15]–[18]. RTK and Precise Point Positioning (PPP) correction services are becoming more capable and more available, giving rise to sub-decimeter convergence in a matter of minutes or even seconds [19]–[22].

Here, we examine the performance of multi-constellation GNSS and Real-Time Kinematic (RTK) corrections on 27,500 km of US and Canadian highways. A representative production-grade automotive multi-constellation GNSS L1-only receiver was driven in tandem with a survey-grade GNSS receiver used as ground truth. Furthermore, the ground truth was an Inertial Navigation System which combines two multi-frequency (L1+L2) GNSS receivers with a tactical-grade Micro-Electro-Mechanical Systems (MEMS) IMU. The two-receiver configuration is used to calibrate heading/attitude and the IMU. The ground truth system utilized networked RTK corrections delivered in real-time via a cellular link.

The analysis presented here is broken down into an examination of GNSS accuracy, availability, and continuity. Position accuracy of the representative production-grade automotive GNSS was determined through direct comparison with the ground truth system. Accuracy of the RTK-enabled ground truth was estimated from the position uncertainty output from the combined INS solution. Availability of satellite visibility and geometry (dilution of precision) was examined for both the survey-grade and production-grade units. Availability of RTK positioning and corrections is also assessed. All accuracy and availability metrics are compared both statistically and geospatially on a map. The probability of continuity loss over timescales associated with critical automotive maneuvers is developed for metrics on satellite visibility, geometry, and RTK positioning. Statistics on outage times are also presented. These performance indicators inform the potential role of GNSS as one of several positioning sensors in achieving the localization requirements of ADAS / V2X / AD features in determining (i) which road, (ii) which lane, or (iii) where in the lane the vehicle is positioned.

## METHODOLOGY

The GNSS data used in this analysis was part of a larger data collection campaign used to develop and validate certain vehicle features. This particular dataset was driven in mid 2018, primarily targeting the population centers of the east and west coasts of the US and Canada on the route shown in Figure 2. Two separate vehicles were used in the data collection, one on each coast. The experiment design and hardware setup were largely the same on both vehicles. A high-level summary of the data collection vehicles and GNSS data collected is given in Figure 3 and Table 2, respectively.

A total of 27,500 kilometers (17,088 miles) was driven, representing 355 hours (~15 full days) of data. This was largely highway driving though some of the route did pass through major urban centers. To put this in perspective, the US National Highway System (NHS) consists of approximately 350,000 kilometers (220,000 miles) of road [23]. Hence, the data shown here represents roughly 8% of the NHS. We feel this data is representative as the route passed through many major urban centers as well as regions of topographic diversity, which represent challenging GNSS environments. To stress the importance of the NHS, though it accounts for only 5% of total road mileage in the United States, the NHS carried 55% of the 1.64 trillion miles travelled in the US in 2015 [24].

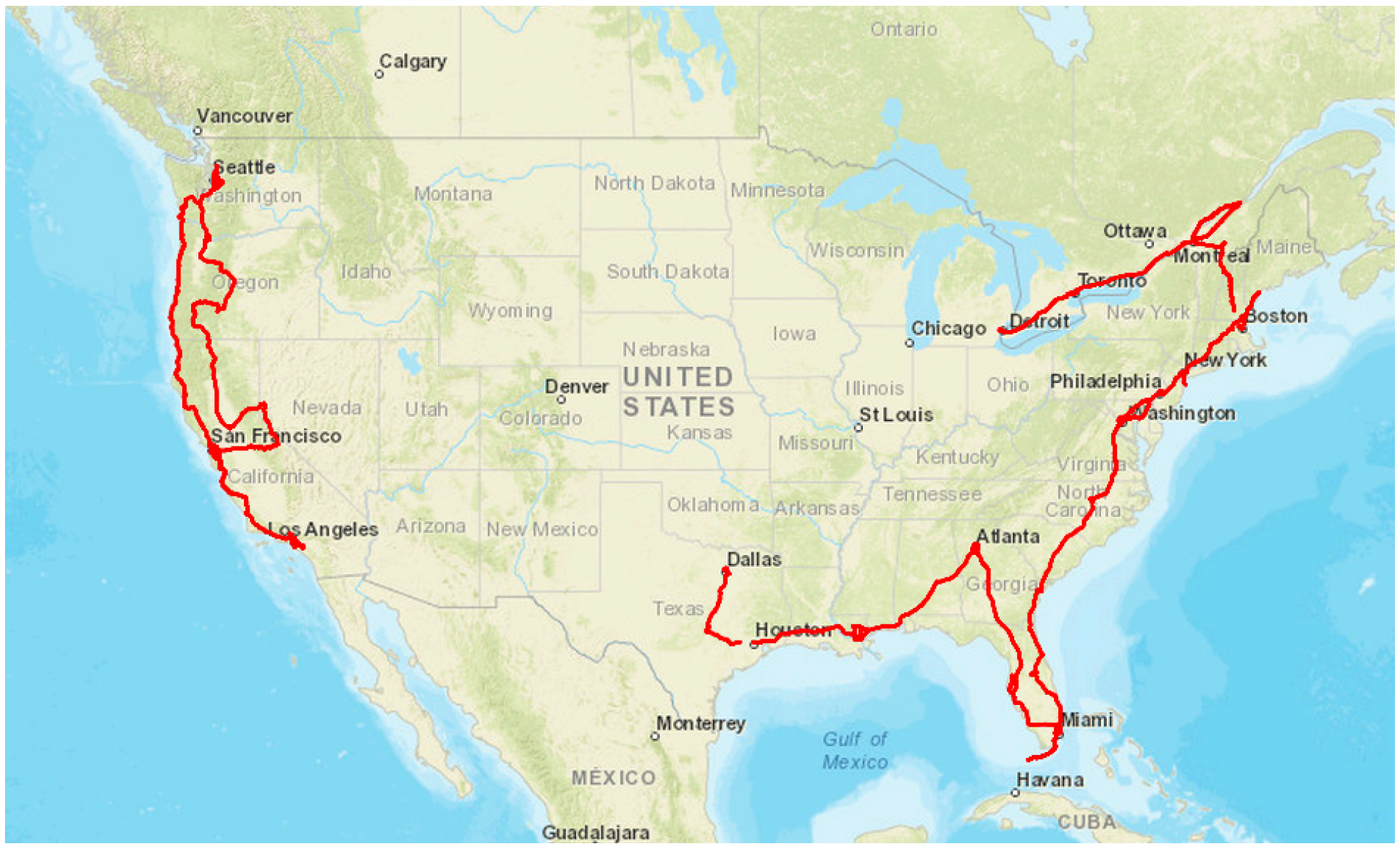

**Figure 2: Map of data collection route.**

Two separate GNSS systems were used in data collection, one representative of that used in production vehicles today and the other a survey-grade system to act as a reference or ground truth. The production-oriented GNSS was automotive grade, multi-constellation capable (GPS + GLONASS + Galileo), and utilized the L1 frequency only. The survey-grade system was an Oxford Technical Solutions (OxTS) RT3000. The RT3000 Inertial Navigation System (INS) combines two survey-grade GNSS receivers with an Inertial Measurement Unit (IMU) to provide the ability to coast through reasonable GNSS outages. The RT3000 units used in this experiment were GPS + GLONASS and capable of tracking both the L1 and L2 frequencies. The MEMS IMU is tactical grade with a gyro bias stability of 2 degrees / hour. The two GNSS receivers are used to constantly calibrate the attitude of the RT3000 IMU, primarily to provide heading.

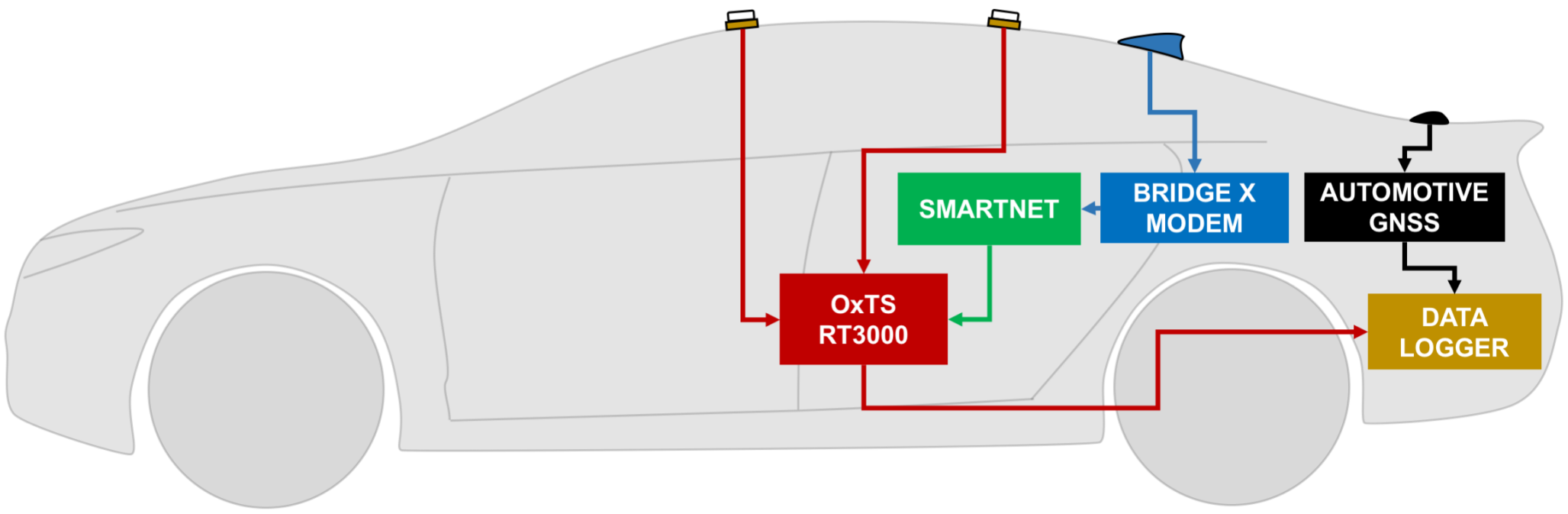

**Figure 3: Experimental vehicle setup. Antenna placement is representative, other equipment layout is spaced out for clarity and may not be representative of physical location on the vehicle.**

**Table 2: Summary of GNSS data collected.**

| Positioning System | Frequencies | Constellations | Signals of Interest | Data Collection Rate [Hz] |
|---|---|---|---|---|
| Automotive GNSS | L1 | GPS GLONASS Galileo | UTC Time, Latitude, Longitude, Altitude, Number of Satellites, HDOP, VDOP, Heading | 1 |
| OxTS RT3000 | L1, L2 | GPS GLONASS | UTC Time, Latitude, Longitude, Altitude, Roll, Pitch, Heading, Number of Satellites, HDOP, PDOP, $1\sigma$ Position Solution Uncertainties, Position mode (RTK Fixed / Float, SPS) Age of Differential Corrections | 30 |

Networked RTK corrections were also used in this experiment. Cellular connectivity was available on the vehicle and corrections were delivered to the RT3000 via an RTK Bridge-X modem from Intuicom Wireless Solutions. The networked RTK corrections were provided by SmartNet, where coverage was available along the entire data collection route.

Both systems provided high level GNSS information including UTC time, latitude, longitude, altitude, number of satellites tracked, and dilution of precision (DOP). The production GNSS data was collected at 1 Hz and the RT3000 at 30 Hz. The RT3000 also output several other parameters related to the full inertial navigation solution including attitude (roll, pitch, heading), position mode (RTK fixed, RTK float, differential code, standard positioning service (SPS), or no service), age of differential corrections, and the $1\sigma$ uncertainty of the position solution. These are summarized in Table 2.

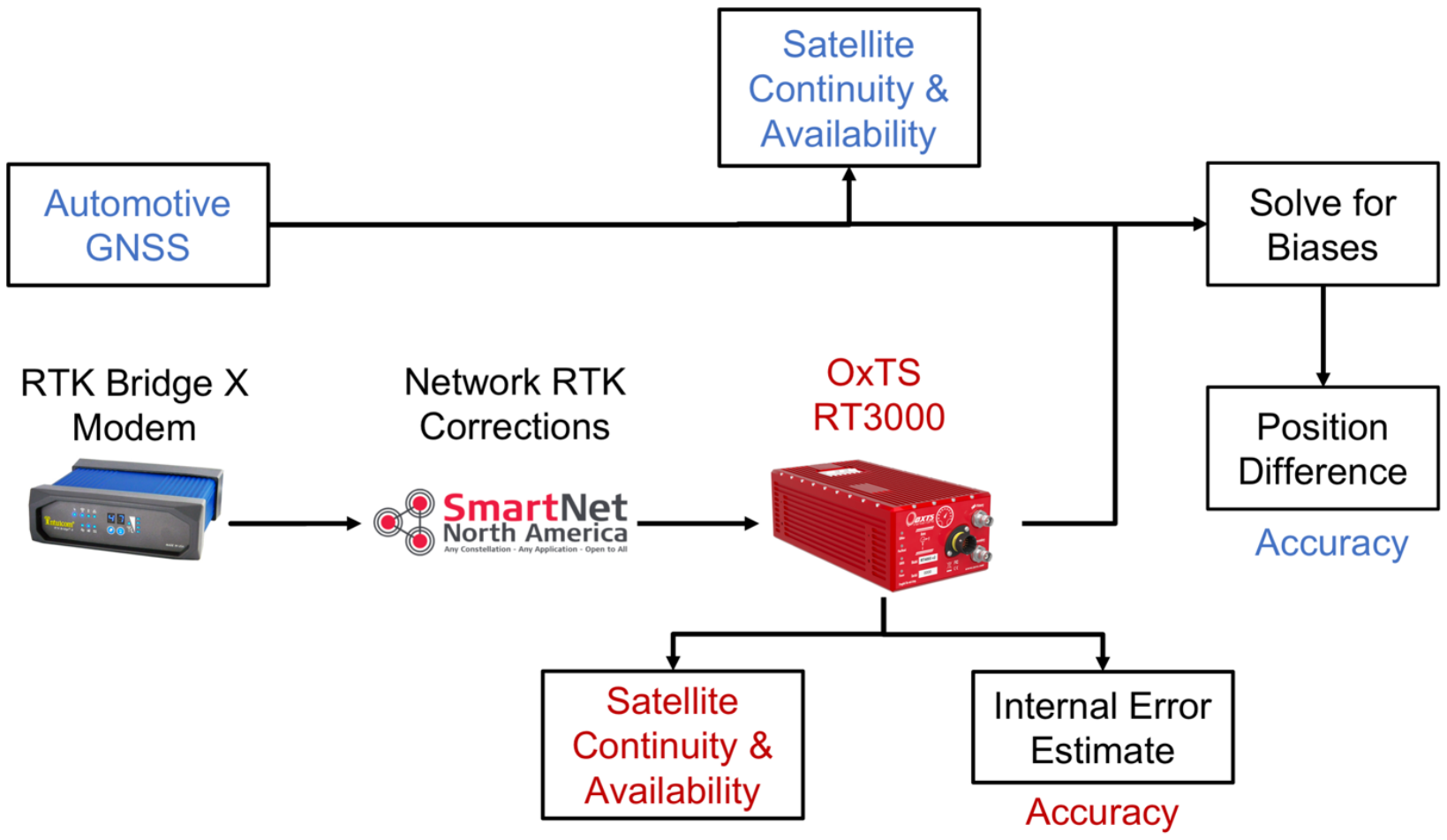

**Figure 4: The data processing pipeline. The RT3000 is used as a reference or ground truth against which to compare the automotive GNSS. Systematic biases in both the body-fixed and global frame are calculated using the methodology given in the appendix before the remaining errors are characterized. Satellites tracked are reported by each system and can be used to assess availability and continuity.**

Using internal tools, time synchronization was achieved between the automotive grade GNSS and RT3000. Aggregate statistics were produced on satellite visibility and geometry to establish availability and continuity. The same was done on parameters related to RTK positioning. Assessment of position accuracy requires a more detailed examination. The production GNSS was compared to what was considered ground truth, the INS solution from the RT3000. The RT3000 is itself not without flaws, hence only highly confident position solutions were used in the comparison. The mathematics of how to compare these systems and remove systematic biases is described in the appendix. These biases are removed as it is assumed they can be reasonably calibrated out as will be described in the next section. This process is summarized in Figure 4.

## RESULTS

This section will be divided into an assessment of on-road GNSS accuracy, availability, and continuity. Accuracy will be broken down by the lateral (side-to-side) and longitudinal (forward-backward) driving directions along with horizontal and vertical for both the stand-alone automotive grade GNSS and survey-grade receiver with RTK corrections (RT3000). Availability will be reported by metrics on satellite visibility, satellite geometry (DOPs), position solution type (RTK fixed, RTK float, standard code phase positioning), and by application type (which road, which lane, where in lane). Continuity is reported by the probability of losing a given number of satellites, level of satellite geometry (DOP), or position type (RTK or otherwise) over time intervals associated with critical vehicle maneuvers, e.g. lane-changes, overtaking, and handover from automated to manual mode.

### *Accuracy*

The production-grade automotive GNSS position error is assessed by direct comparison to the RT3000 which is considered ground truth using the methodology described in the appendix. To add reliability to this assertion, only highly confident RT3000 INS position solution points were used in the comparison. These were taken as having a horizontal position uncertainty of $\sigma_H < 10$ centimeters, a parameter output from the RT3000 itself. These were primarily the RTK-fixed positions of the RT3000 and hence the most reliable. As will be shown, this is taking points that are at least an order of magnitude more accurate than the automotive GNSS under investigation. The error distribution of the RT3000 itself was estimated from this reported uncertainty output $\sigma_H$.

**Table 3: Summary of the automotive GNSS and the RT3000 position accuracy. The automotive GNSS accuracy is assessed through comparison with the RT3000. The RT3000 accuracy is estimated as its reported position uncertainty.**

| Positioning System | Lateral [m] | | | Longitudinal [m] | | | Horizontal [m] | | | Vertical [m] | | |
|---|---|---|---|---|---|---|---|---|---|---|---|---|
| | 68% | 95% | 99% | 68% | 95% | 99% | 68% | 95% | 99% | 68% | 95% | 99% |
| Automotive GNSS | 1.92 | 3.88 | 5.74 | 2.11 | 4.44 | 7.95 | 3.07 | 5.30 | 9.38 | 4.59 | 9.42 | 12.83 |
| OxTS RT3000 | 0.18 | 0.73 | 2.60 | 0.18 | 0.75 | 2.88 | 0.26 | 1.05 | 3.91 | 0.31 | 1.34 | 2.56 |

The cumulative position error distributions for the automotive GNSS and RT3000 are given in Figures 5 and 6, respectively, and a high-level summary is given in Table 3. These show the horizontal and vertical position error, where horizontal is further broken down into lateral and longitudinal components.

Many factors contribute to the performance in these directions including the vehicle-relative location of obstructions due to buildings, trees, signs, and other vehicles, affecting satellite geometry and multipath. Results indicate longitudinal accuracy to be slightly worse compared to lateral for the automotive GNSS. This result could also be indicative of delays in the system where at typical highway speeds, the 68$^{th}$ percentile difference of 20 centimeters between lateral and longitudinal could be the manifestation of a time mismatch of 5 milliseconds, not unreasonable for vehicle Controller Area Network (CAN) signals [25], [26].

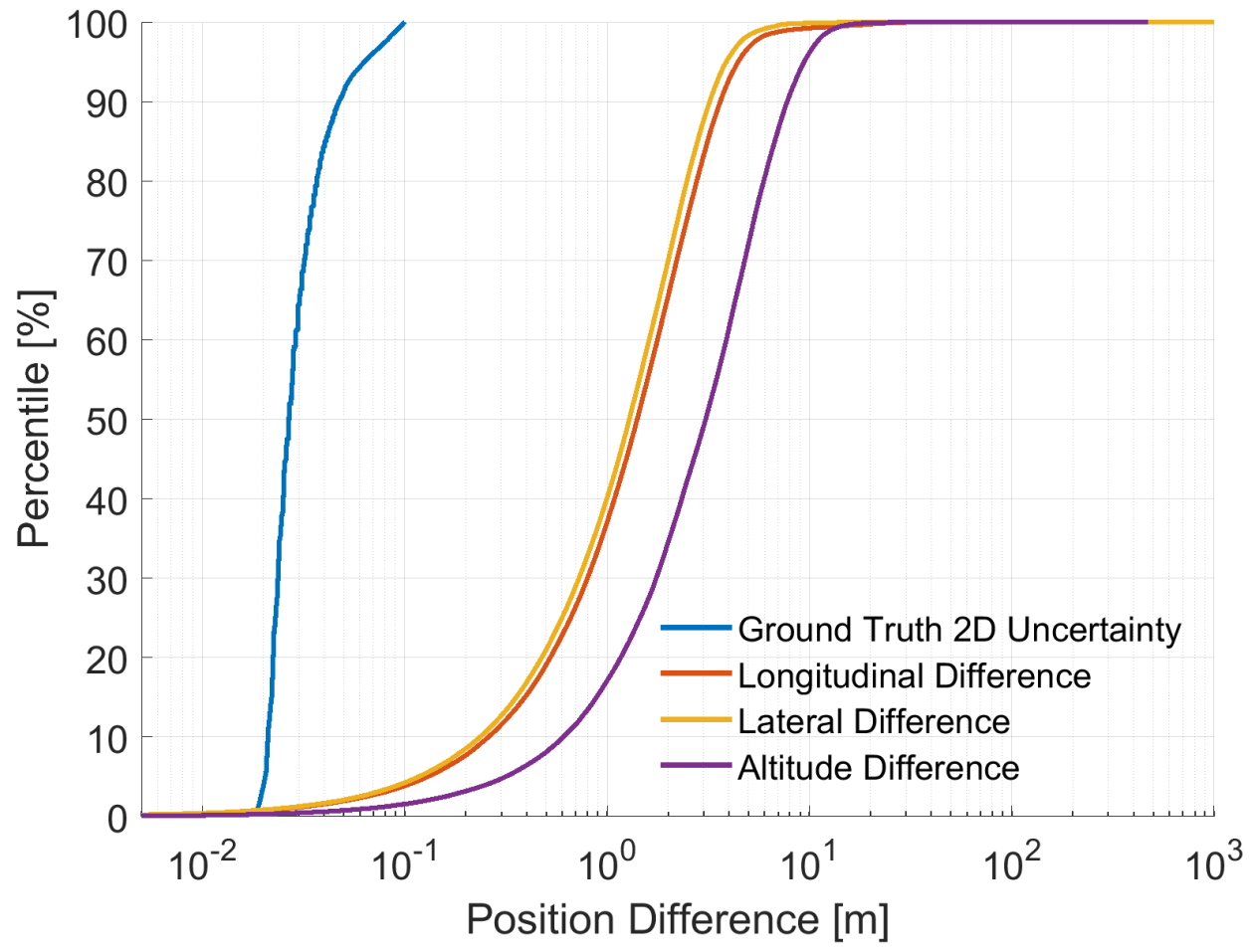

**Figure 5: Empirical cumulative distribution of the position difference between the automotive GNSS and ground truth (RT3000). This used only highly confident ground truth points in the comparison.**

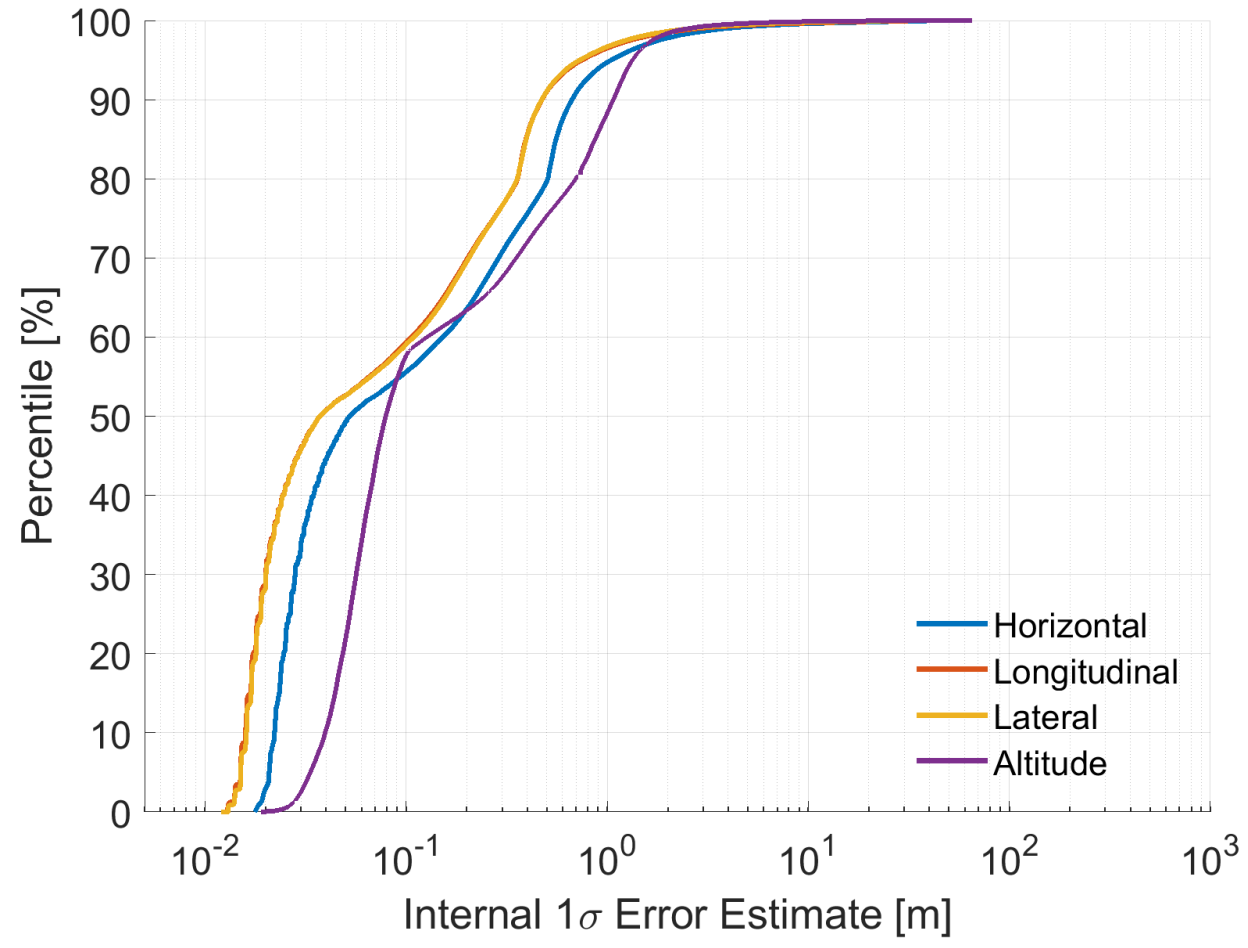

**Figure 6: Empirical cumulative distribution of the ground truth (RT3000) reported 1σ error estimates over the entire route.**

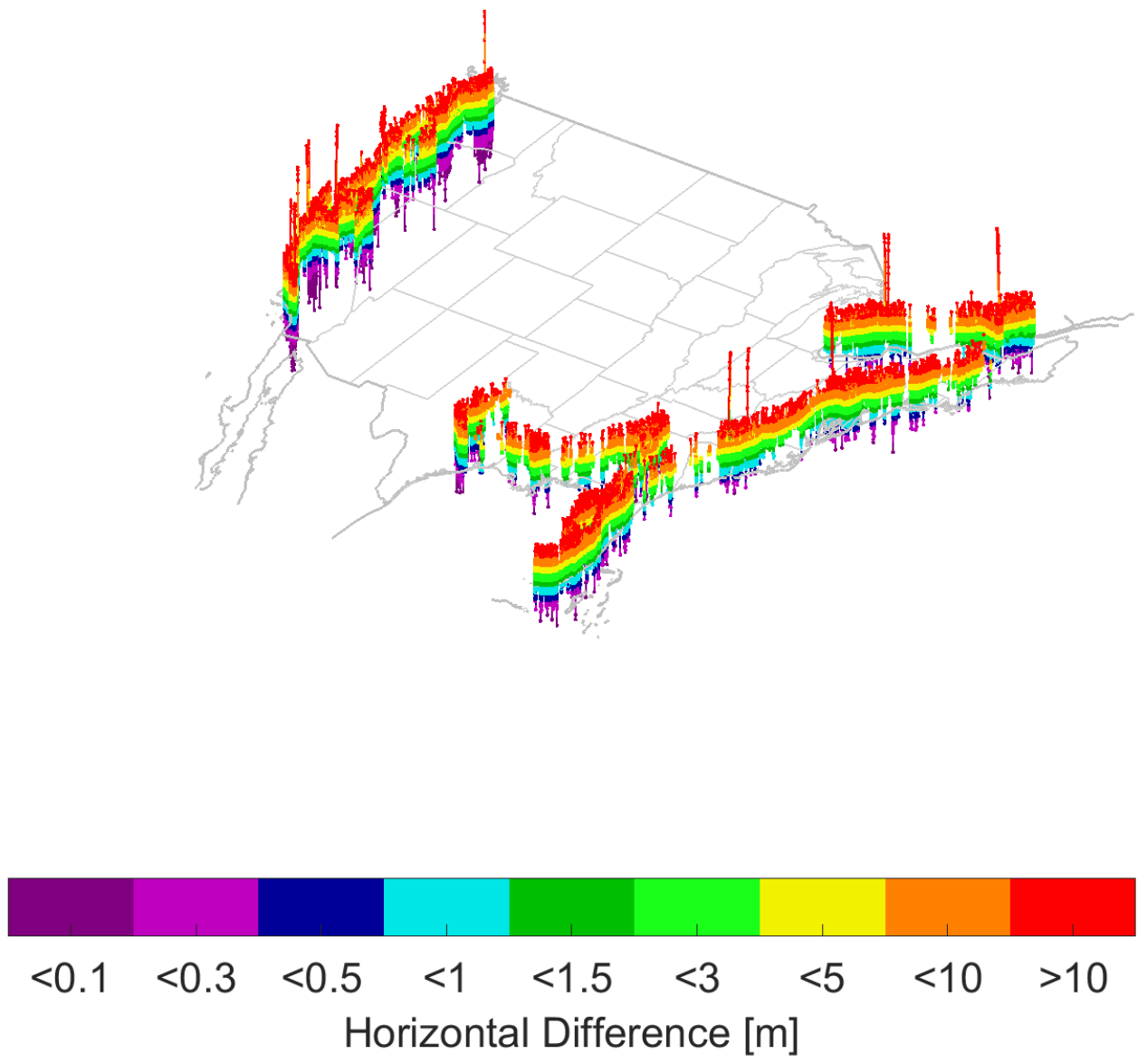

**Figure 7: Geospatial distribution of the position difference between the automotive GNSS and ground truth RT3000. This used only highly confident ground truth points in the comparison.**

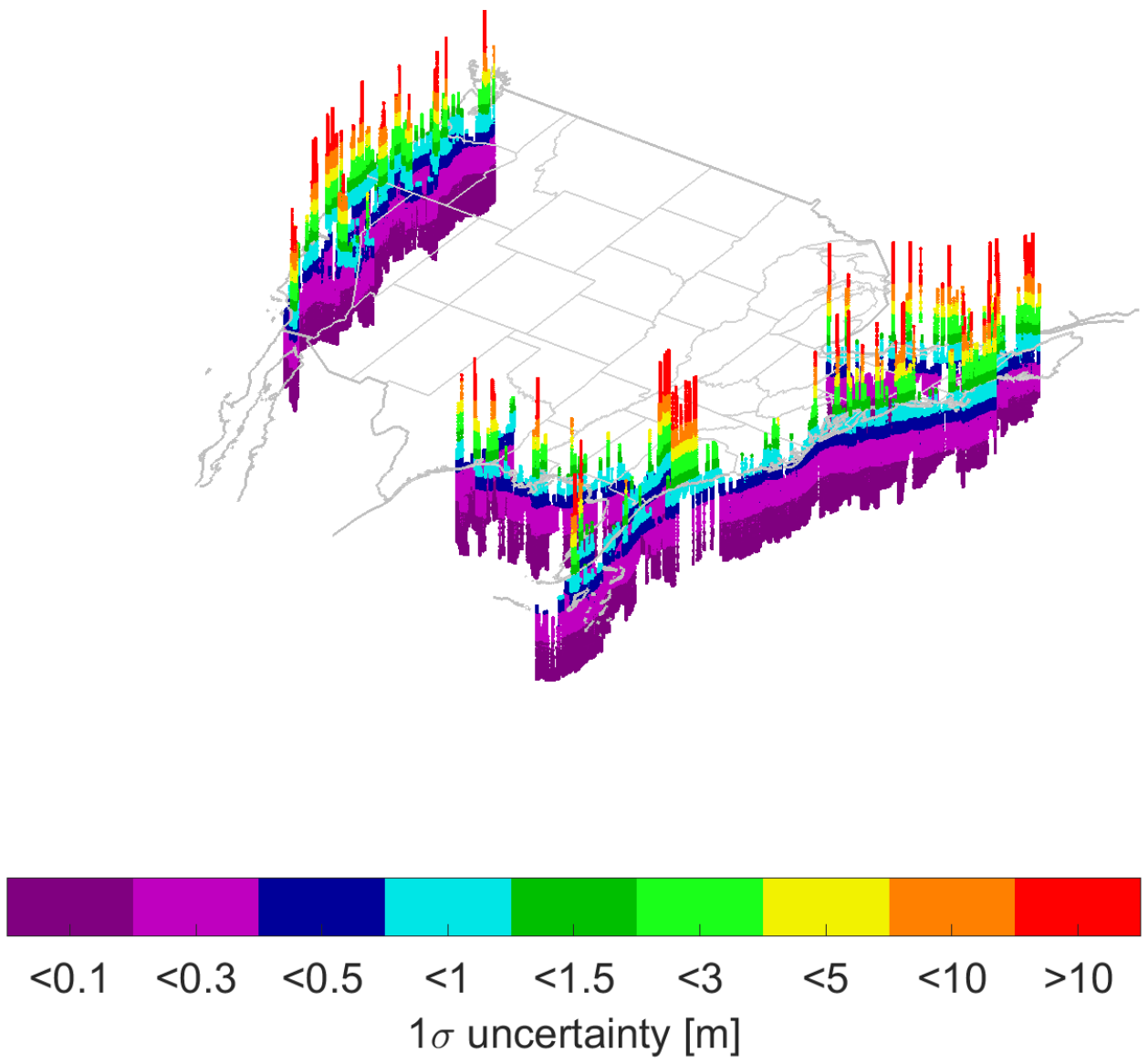

**Figure 8: Geospatial distribution of the ground truth (RT3000) reported 1-σ horizontal error estimate.**

The RT3000 cumulative position error distribution in Figure 6 shows no significant difference between the lateral and longitudinal position errors. Closer examination shows that this distribution has three distinct regimes corresponding to the quality of the different positioning modes. In the centimeter regime, there are the 'RTK fixed' solutions, where the system is successful in carrier phase integer ambiguity resolution. In the decimeter regime, there are the 'RTK float' solutions, where only floating-point estimates of the carrier phase integer ambiguities are available. The final meter-level regime is standard code phase positioning where either RTK corrections were unavailable or obstructions disrupted carrier phase tracking. Furthermore, this shows that only 55% of positions were estimated as better than 10 centimeters horizontal. These represent predominantly RTK fixed solutions and were those used in the comparison with the automotive GNSS. Hence, the full distribution shown for the RT represents both good and bad GNSS conditions. In contrast, the automotive GNSS error is estimated in places where the RT3000 showed good performance and was likely in favorable GNSS environments.

The geospatial distribution of horizontal position errors for the automotive GNSS and RT3000 are given in Figures 7 and 8, respectively. This shows the spatial uniformity of the errors. Though there are some spikes in Figure 7 with the automotive GNSS, these locations correspond to areas that pass through major urban centers and hence urban canyons with poor satellite visibility. In these environments, poor geometry is further compounded by the presence of multipath and Non-Line-Of-Sight (NLOS) signals. The RT3000 shows a higher spatial frequency of spikes. Though the magnitude of these spikes is typically smaller than with the production GNSS, their increased frequency can be attributed to the fact that the automotive GNSS was evaluated on a subset of the RT3000 positions in places the RT3000 was highly confident, likely representing favorable GNSS environments.

### *Availability*

This section examines the availability of position service level in terms of (i) which road, (ii) which lane, and (iii) where in lane performance along with satellite visibility and geometry. Elements specific to RTK positioning and corrections will be discussed later in a devoted section.

**Table 4: Position service level availability for the automotive GNSS and RT3000.**

| Position Service Level | Lateral Accuracy [m] | Availability [%] | |
|---|---|---|---|
| | | Automotive GNSS | OxTS RT3000 |
| Which Road | < 5 | 98.3 | 99.5 |
| Which Lane | < 1.5 | 56.7 | 98.1 |
| Where in Lane | < 0.5 (Highway) | 21.0 | 91.1 |
| | < 0.3 (Local Streets) | 12.7 | 76.6 |

Table 4 shows position service level availability broken down by (i) which road, (ii) which lane, and (iii) where in lane performance. This indicates that the automotive GNSS is capable of road determination 98.3% of the time, though it is less suitable for lane-level or where-in-lane applications at 56.7% and 21.0%, respectively. By comparison, the RT3000 is capable of road determination 99.5% of the time, lane determination at 98.1%, and where-in-lane highway applications at 91.1%. It should again be noted that at the points where the automotive GNSS was evaluated, the RT3000 was confident at <10 cm, hence having 100% availability in all of the categories listed at these epochs. Something not considered here is the error budget allocation needed in the global accuracy of the map in these use cases. Ultimately, if the desire is to perform lane determination or lane-keeping through a combination of GNSS with a lane-level map, then the global uncertainty of the map and the uncertainty of the GNSS position stack up. This will ultimately lead to more stringent requirements on GNSS; however, we feel these numbers are a good starting point.

At the root of positioning performance is the availability of satellites both in numbers and spatial diversity (i.e. geometry). In this context, this is a combined measure of satellites available above the horizon and environmental factors causing satellite obstructions. A summary of these results is given in Table 5.

**Table 5: Summary of reported satellite geometry and satellite visibility for the automotive GNSS and RT3000.**

| Positioning System | HDOP | | | Number of Satellites (Least) | | |
|---|---|---|---|---|---|---|
| | 68% | 95% | 99% | 68% | 95% | 99% |
| Automotive GNSS | 0.60 | 0.80 | 0.80 | 16 | 13 | 12 |
| OxTS RT3000 | 1.30 | 3.70 | 11.20 | 8 | 4 | 0 |

The cumulative distribution of reported satellite visibility for the automotive GNSS and RT3000 are given in Figures 9 and 10, respectively. This shows the 68$^{th}$ percentile number of satellites available for the automotive GNSS to be 16 with GPS + GLONASS + Galileo compared to 8 with the GPS + GLONASS RT3000. The 99$^{th}$ percentile shows an even starker difference at 12 and 0, respectively. In part, this signifies the increased availability from a third constellation, adding approximately 8 more satellites above the horizon. However, the full story is given by the geospatial distribution of satellite visibility for the automotive GNSS and RT3000 shown in Figures 11 and 12. This shows a difference in how satellite visibility is reported between these two receivers. The RT3000 is constantly dropping down to zero as a result of overpasses and other occlusions. The automotive GNSS does not show this behavior but instead gives more of what appears to be an average sense of satellites above the horizon rather than satellites instantaneously tracked. The RT3000 shows spatial uniformity in that good and bad satellite visibility is seemingly evenly dispersed. The automotive GNSS has low points only in a handful of places which correspond to major urban centers with prolonged satellite outages.

Satellite geometry as described by Horizontal Dilution of Precision (HDOP) is also summarized in Table 5. The cumulative reported HDOP for the automotive GNSS and RT3000 are given in Figures 13 and 14, respectively. Similarly, the geospatial distributions are given in Figures 15 and 16. These again show spatial uniformity. Though there are some spikes with the automotive GNSS, these locations again correspond to areas that pass through major urban centers and hence urban canyons, showing a strong correlation between satellites in view and HDOP. Another observation is that the west coast generally has slightly worse HDOP overall. This characteristic could be explained by the generally more mountainous terrain, which can introduce a higher elevation angle mask. The RT3000 again shows HDOP to be spatially changing at a much higher frequency, indicating similar behavior as with satellite visibility.

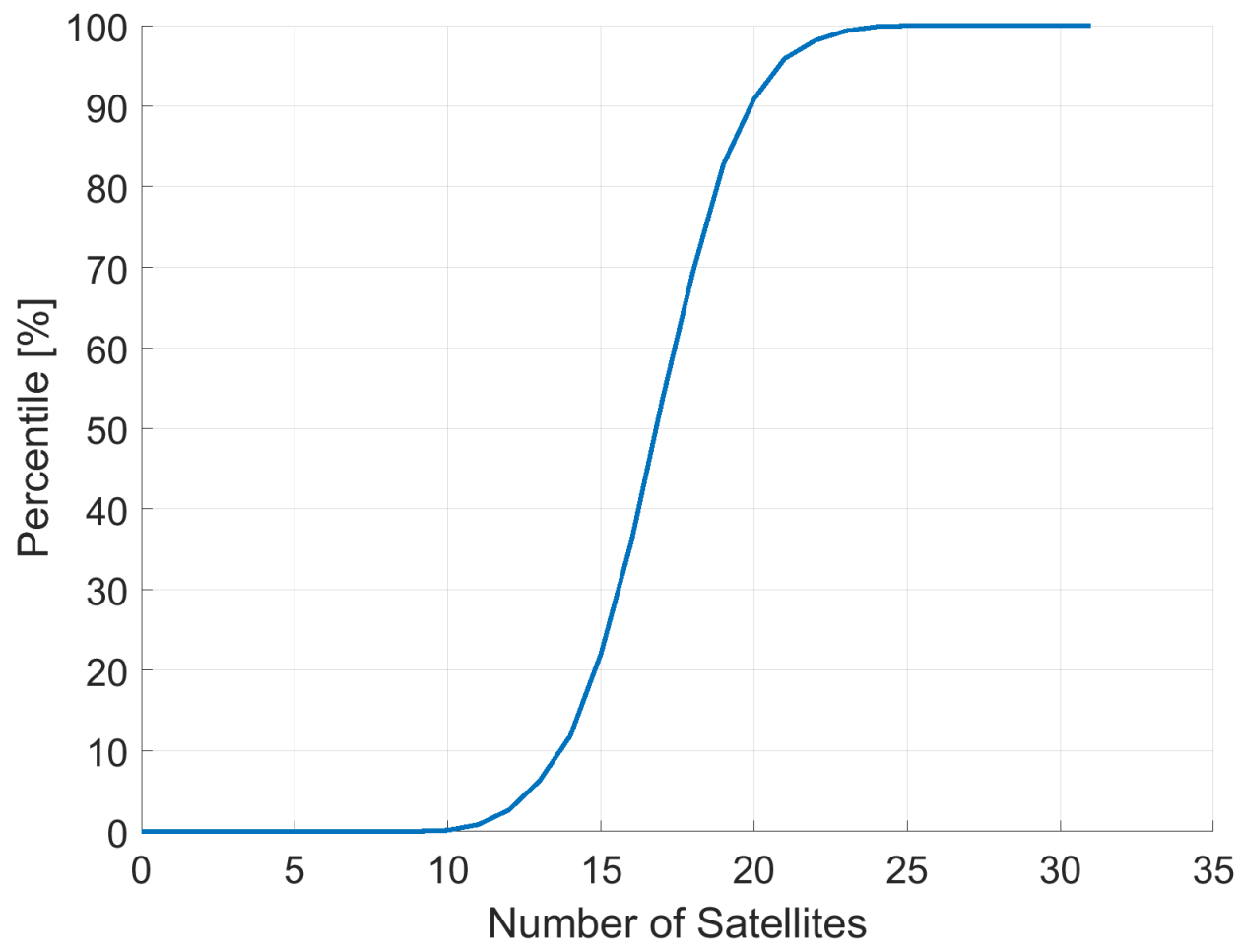

**Figure 9: Empirical cumulative distribution of reported satellite visibility for the automotive GNSS.**

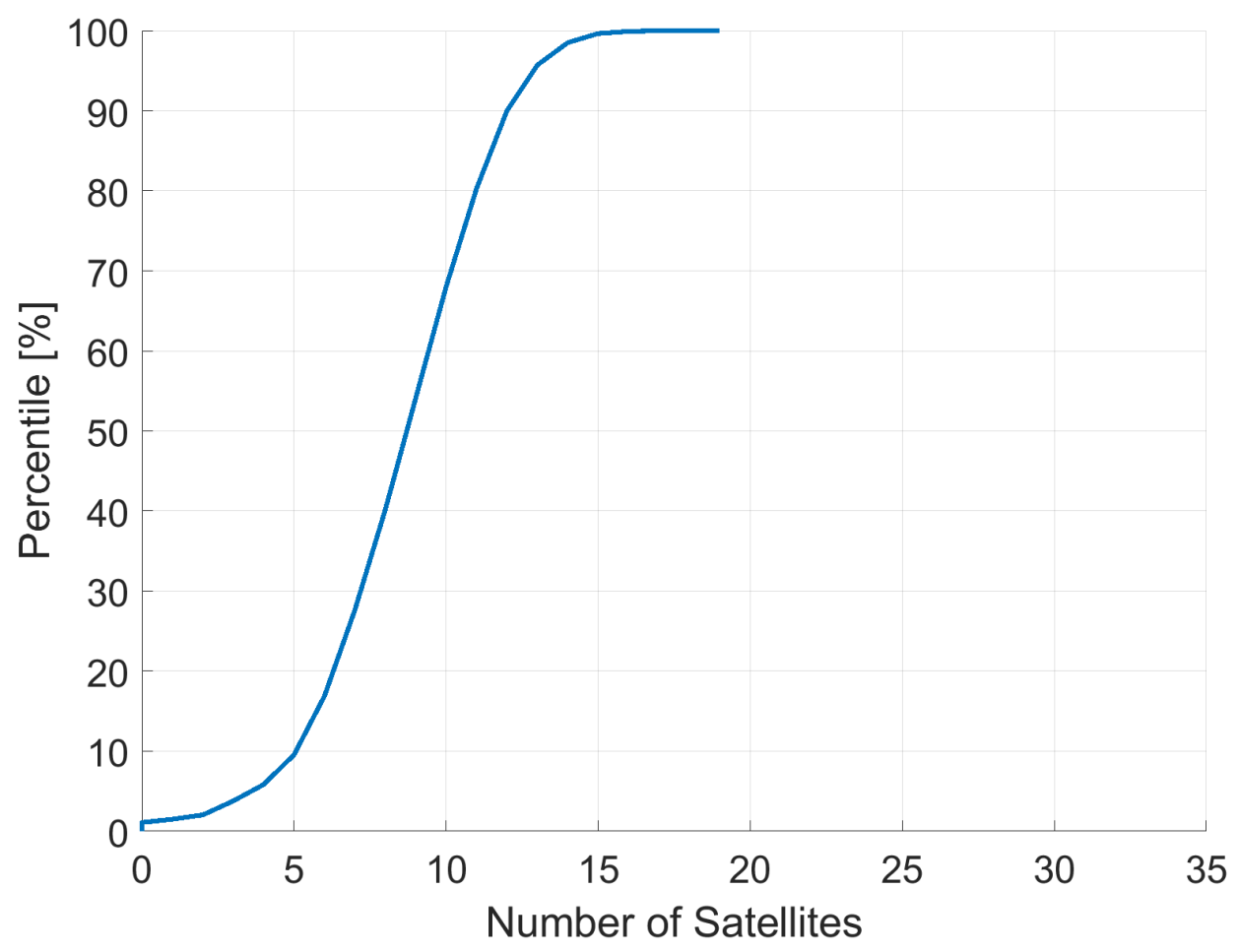

**Figure 10: Empirical cumulative distribution of reported satellite visibility for the RT3000.**

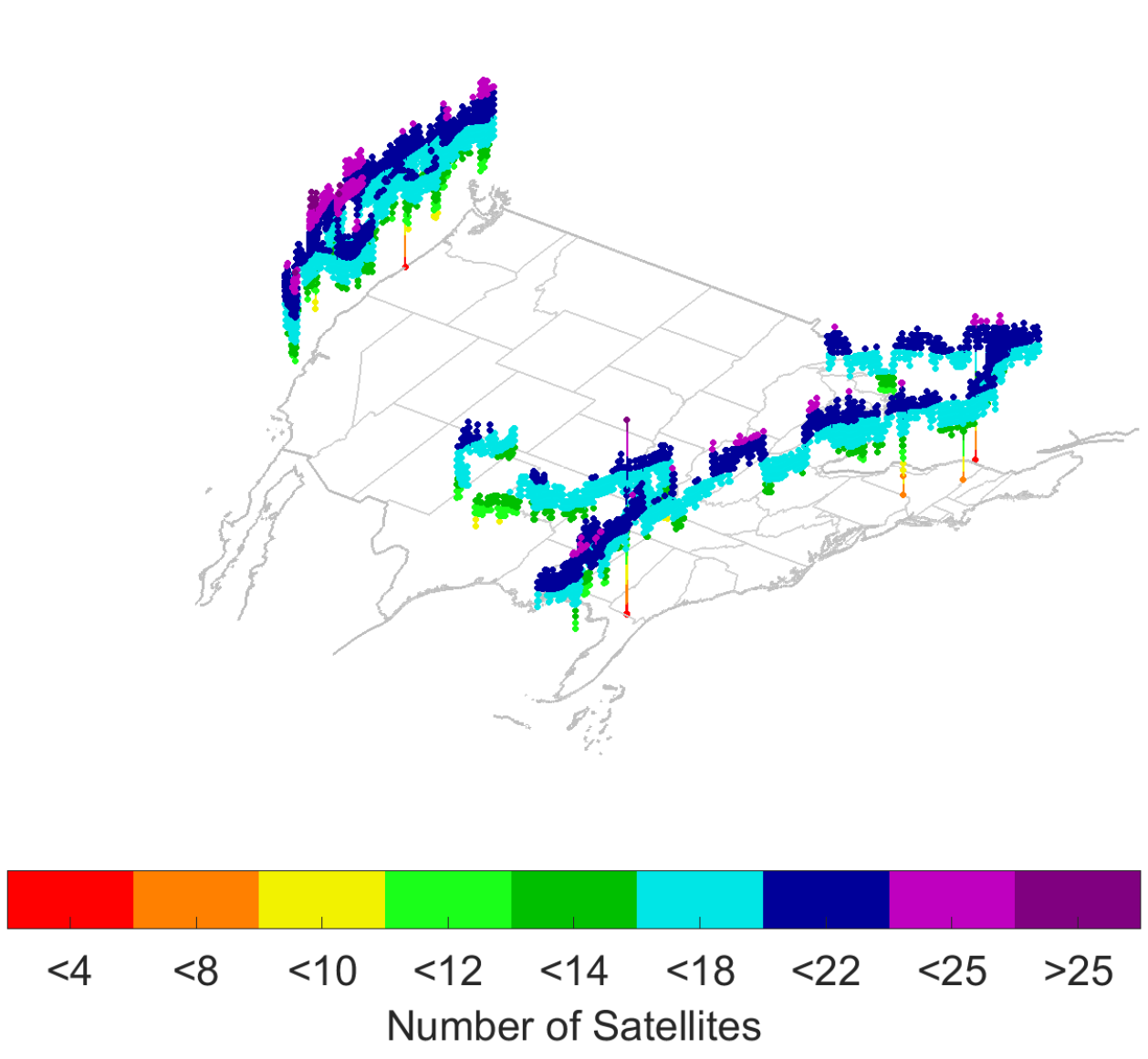

**Figure 11: Geospatial distribution of reported satellite visibility for the automotive GNSS.**

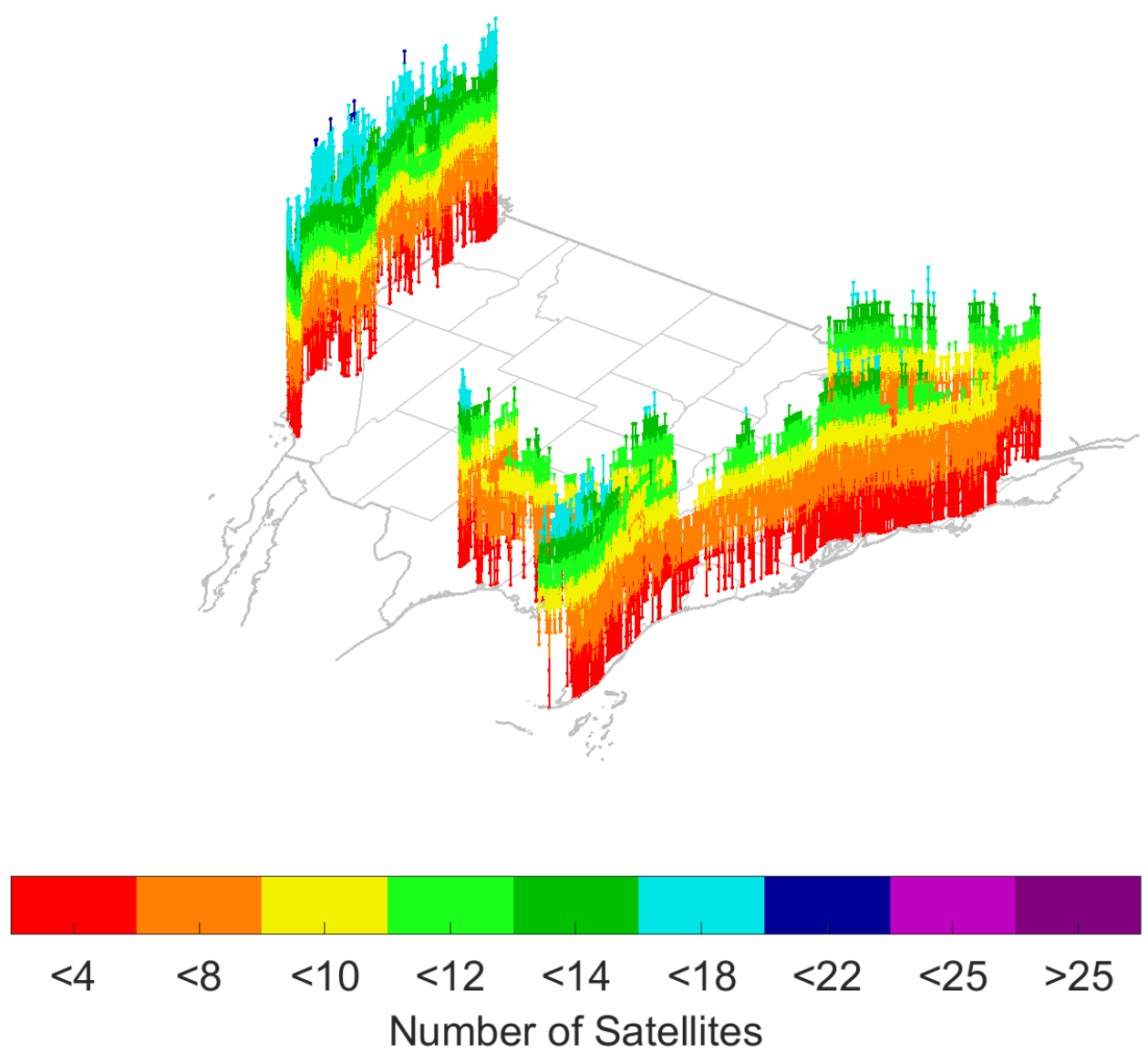

**Figure 12: Geospatial distribution of reported satellite visibility for the RT3000.**

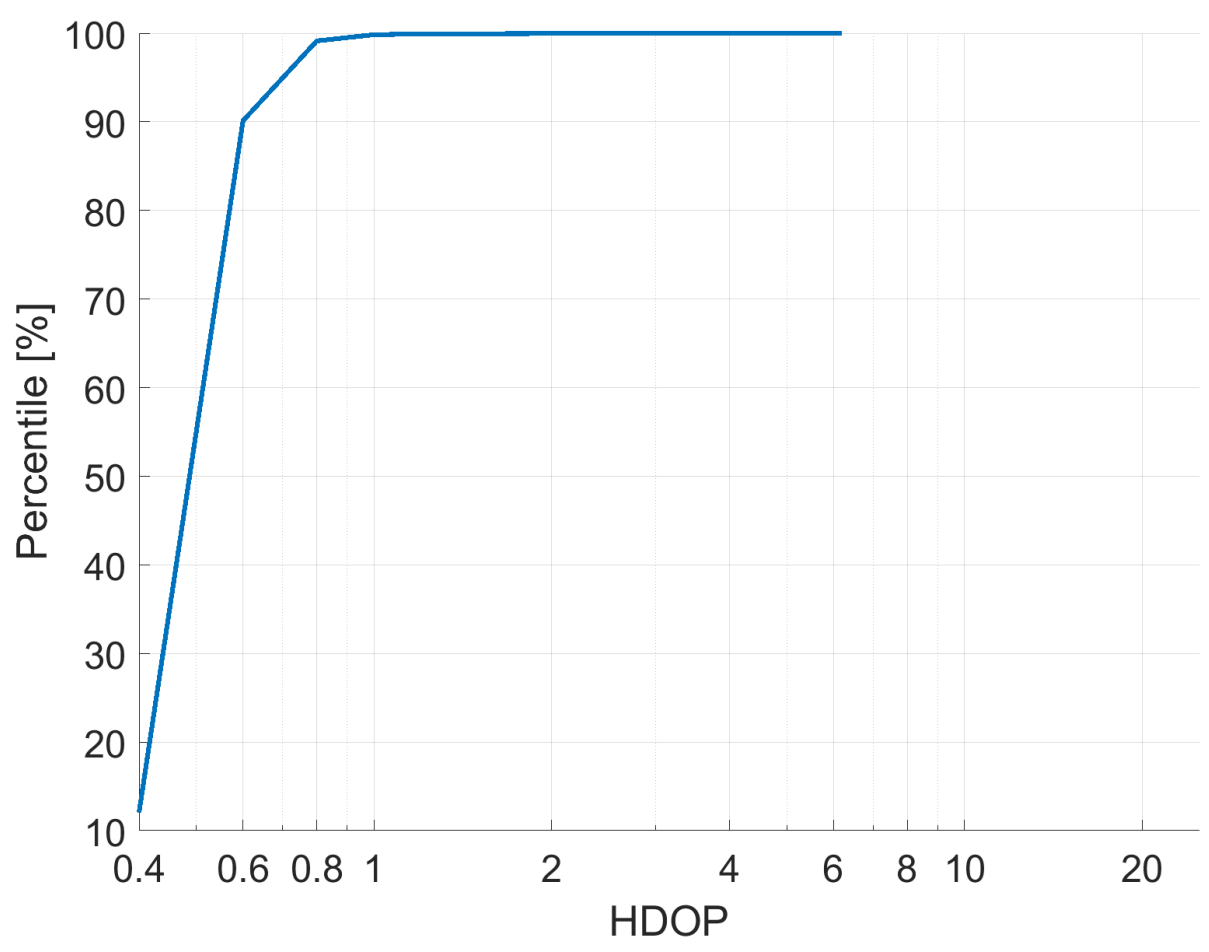

**Figure 13: Empirical cumulative distribution of reported horizontal dilution of precision for the automotive GNSS.**

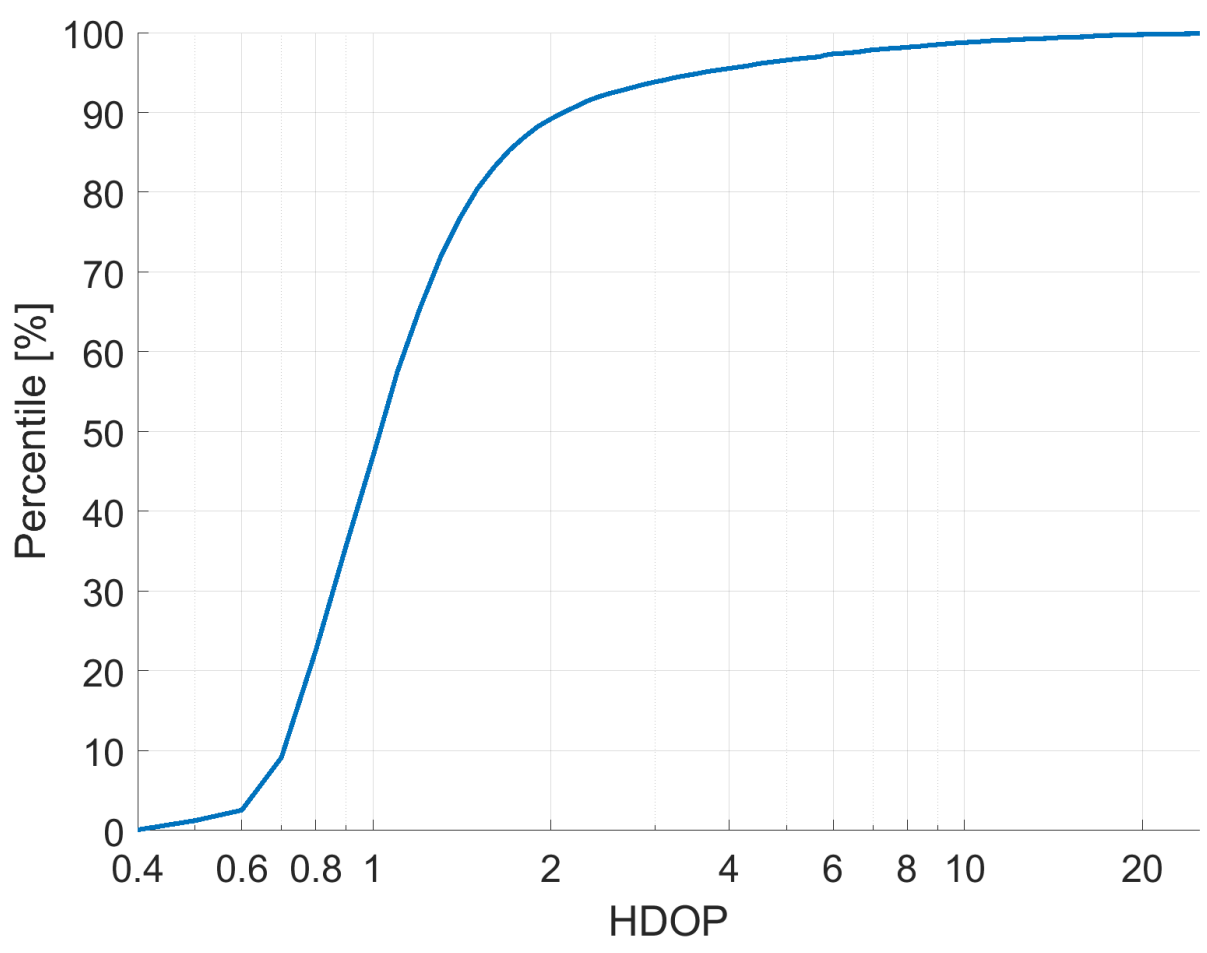

**Figure 14: Empirical cumulative distribution of reported horizontal dilution of precision for the RT3000.**

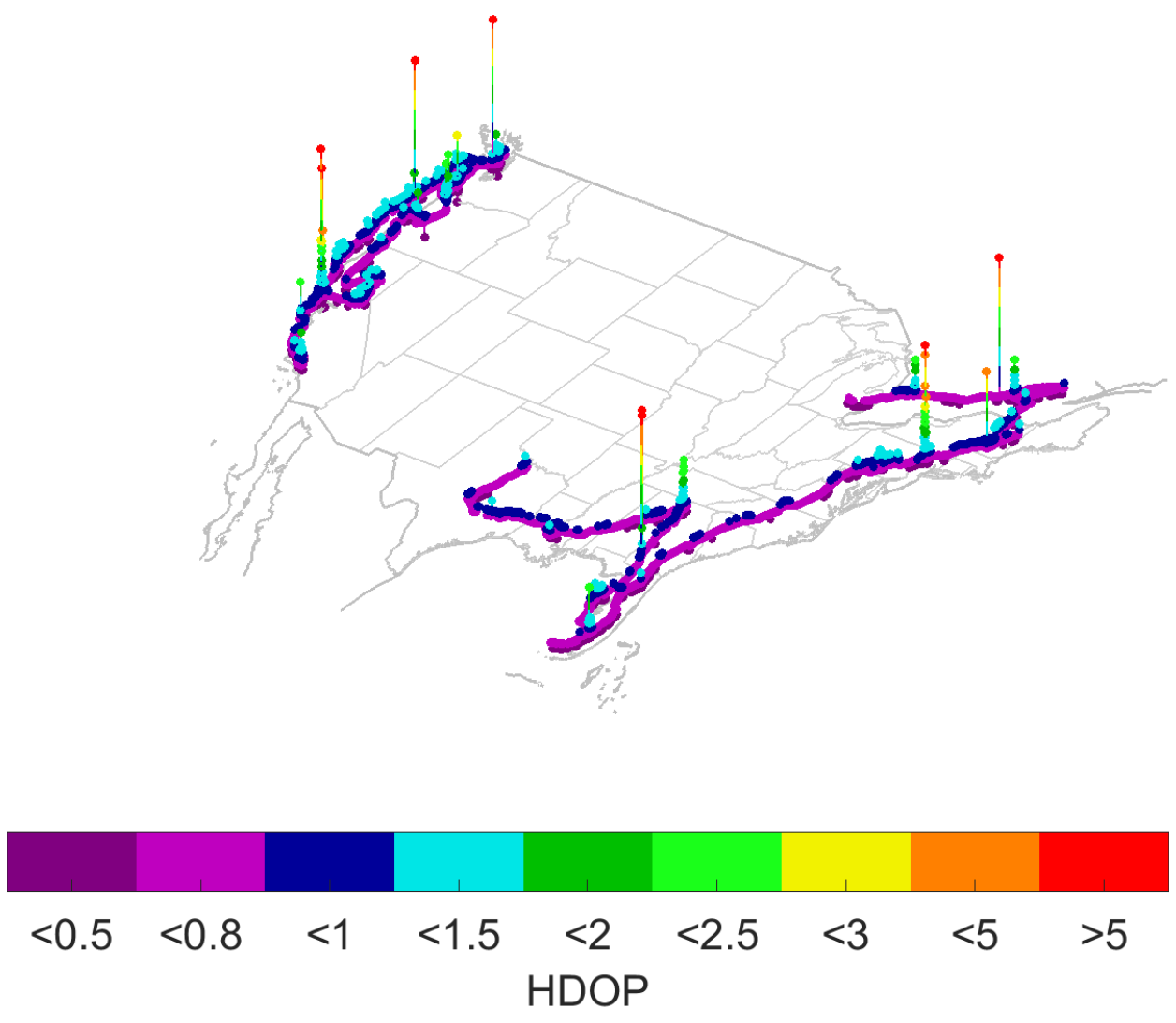

**Figure 15: Geospatial distribution of reported horizontal dilution of precision for the automotive GNSS.**

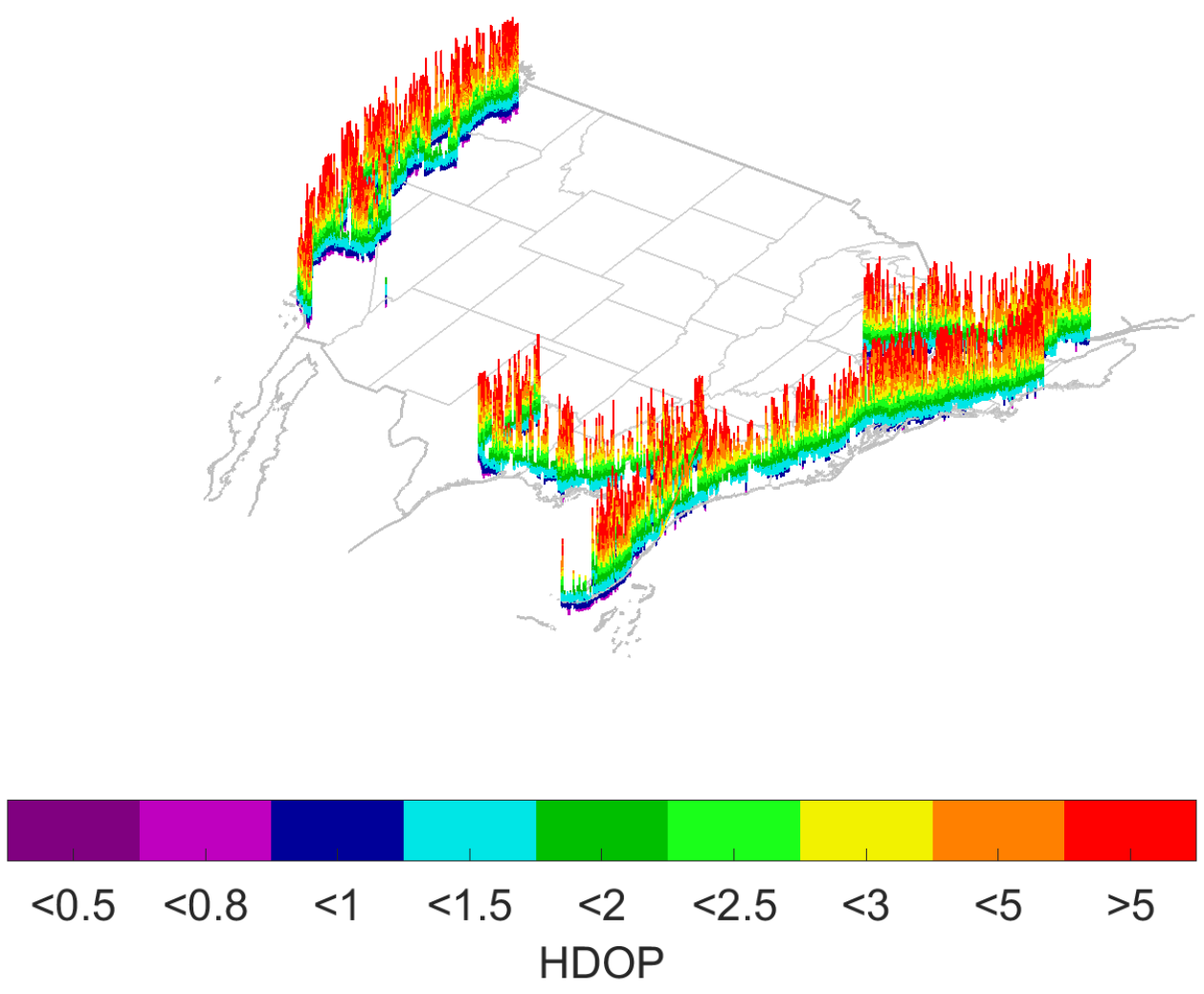

**Figure 16: Geospatial distribution of reported horizontal dilution of precision for the RT3000.**

### *Continuity & Outage Times*

In this section, the continuity of satellite visibility and geometry is examined. Strictly speaking in the context of GNSS, continuity is defined as the likelihood that the navigation system supports the accuracy and integrity requirements for the duration of the intended operation [27], [28]. Since the accuracy and integrity requirements have not been strictly defined in this context, this section will focus on satellite continuity in terms of number and spatial diversity. Furthermore, the emphasis will be on the RT3000 as its data was available at a high rate (>30 Hz) compared to the automotive GNSS (1 Hz), consequently this represents GPS + GLONASS. As was shown in the previous section on availability, the reported satellite visibility and dilution of precision from the automotive GNSS was more indicative of satellites in view in an average sense, not what was instantaneously tracked. The continuity of RTK positioning will be discussed in the section on RTK performance.

In critical aviation maneuvers, such as precision approach or instrument landing, the timescale of interest for continuity is 15 seconds [27]. The allocated continuity loss risk over this time period is $8\times10^{-6}$ for both precision approach and instrument landing [27]. Assumptions with this allotment are that GNSS is the primary navigation signal and that the sky is unobstructed and hence loss in continuity is coming from the signal in space.

In the automotive domain, the timescales of interest correspond to the duration of critical maneuvers such as lane changes, overtaking, and in the case of highly automated systems, driver handover time. Toledo and Zohar [29] show the duration of lane changes to typically be 4 seconds but could take up to 13 seconds in some conditions. Vlahogianni [30] showed the duration of overtaking maneuvers on two-lane highways to typically be 7 seconds but could take up to 17 seconds. Automated to manual driving handover takes between 2.8 and 23.8 seconds based on a study presented by Eriksson and Stanton [31], assuming the driver is involved in a secondary task such as reading before regaining control.

Based on these maneuvers, the probability of continuity loss in 4, 7, 15, and 30 second intervals was examined. These are summarized in Table 6. A more detailed look at satellite visibility and HDOP is given in Figures 17 and 18, respectively. These show the spread of continuity loss to be small over the time intervals of interest for a given number of satellites or level of HDOP. Furthermore, notice that even for the shortest interval of 4 seconds, probabilities of continuity loss are on the order of $10^{-1}$, nearly 5 orders of magnitude worse than numbers assumed in critical aviation maneuvers. These directly impact the continuity of the position solution which will be shown to be on the same order in the next section.

**Table 6: Summary of satellite visibility and satellite geometry probability of continuity loss.**

| | Number of Satellites | | | | | HDOP | | | | |
|---|---|---|---|---|---|---|---|---|---|---|
| Interval [sec] | 4 | 6 | 8 | 10 | 12 | 0.6 | 1.0 | 1.5 | 3.0 | 5.0 |
| 4 | $5.6\times10^{-2}$ | $1.3\times10^{-1}$ | $3.0\times10^{-1}$ | $5.1\times10^{-1}$ | $7.0\times10^{-1}$ | $9.9\times10^{-1}$ | $5.8\times10^{-1}$ | $2.4\times10^{-1}$ | $7.8\times10^{-2}$ | $4.6\times10^{-2}$ |
| 7 | $6.6\times10^{-2}$ | $1.5\times10^{-1}$ | $3.3\times10^{-1}$ | $5.3\times10^{-1}$ | $7.2\times10^{-1}$ | $9.9\times10^{-1}$ | $6.0\times10^{-1}$ | $2.6\times10^{-1}$ | $9.0\times10^{-2}$ | $5.4\times10^{-2}$ |
| 15 | $9.2\times10^{-2}$ | $2.0\times10^{-1}$ | $3.9\times10^{-1}$ | $5.8\times10^{-1}$ | $7.5\times10^{-1}$ | $9.9\times10^{-1}$ | $6.5\times10^{-1}$ | $3.1\times10^{-1}$ | $1.2\times10^{-1}$ | $7.5\times10^{-2}$ |
| 30 | $1.3\times10^{-1}$ | $2.6\times10^{-1}$ | $4.6\times10^{-1}$ | $6.4\times10^{-1}$ | $7.8\times10^{-1}$ | $9.9\times10^{-1}$ | $7.0\times10^{-1}$ | $3.8\times10^{-1}$ | $1.6\times10^{-1}$ | $1.1\times10^{-1}$ |

The next component is characterization of outage times. The cumulative distribution of outage times for satellite visibility and HDOP is given in Figures 19 and 20. For a threshold of 6 satellites in view, this shows the median outage time to be 1.5 seconds and the 95$^{th}$ percentile to be 26 seconds. This will be examined in more detail in the next section on RTK performance.

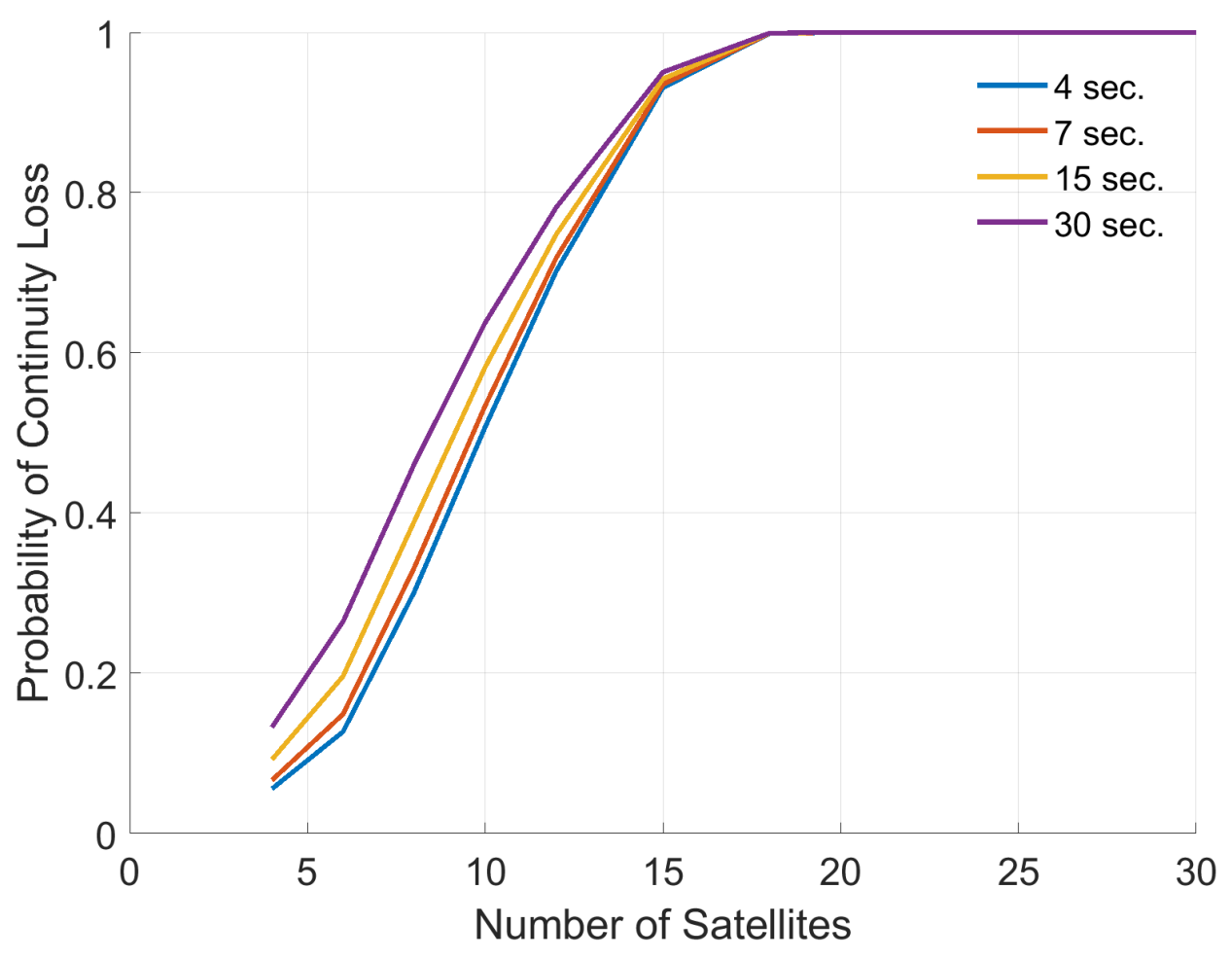

Figure 17: Empirical satellite visibility probability of continuity loss for the RT3000.

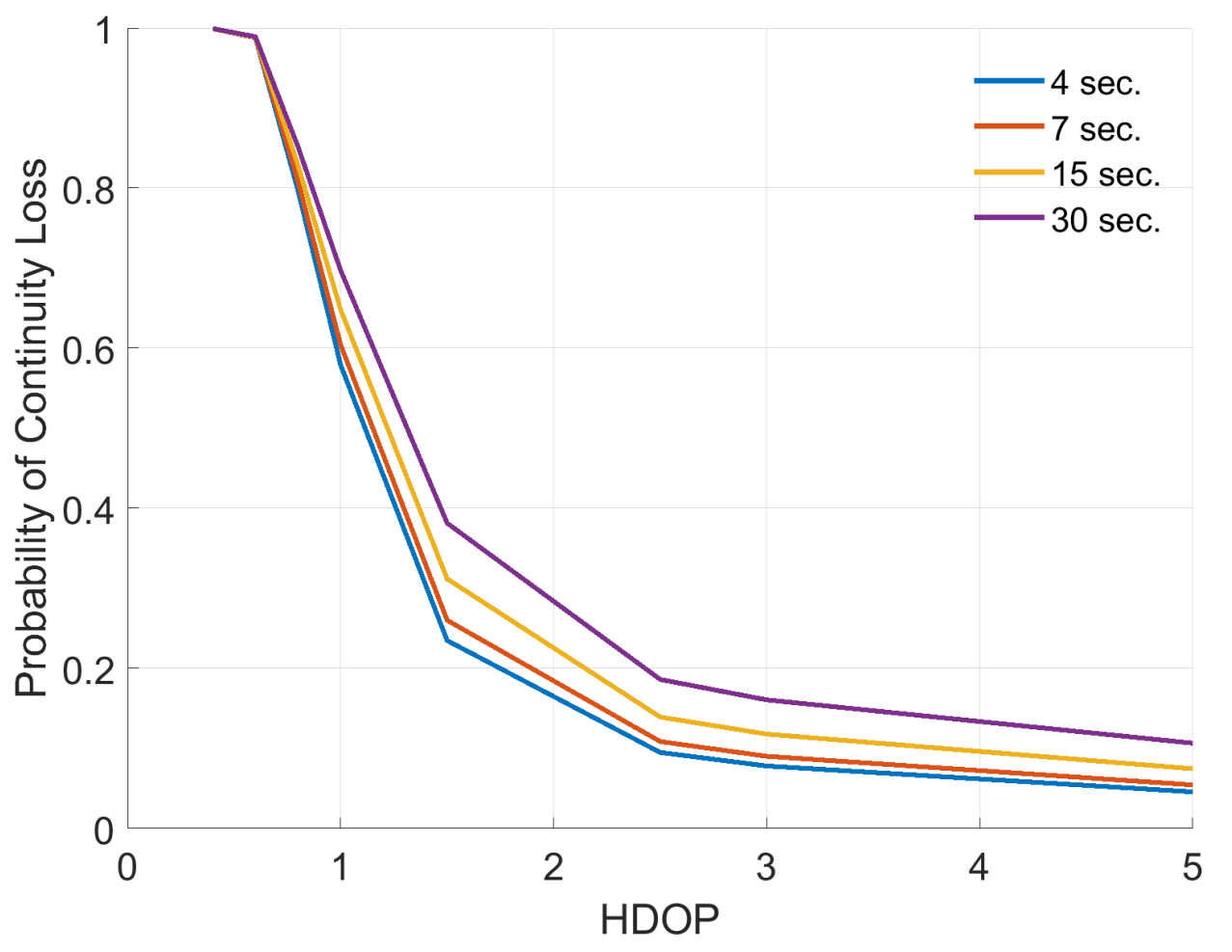

Figure 18: Empirical satellite geometry (HDOP) probability of continuity loss the RT3000.

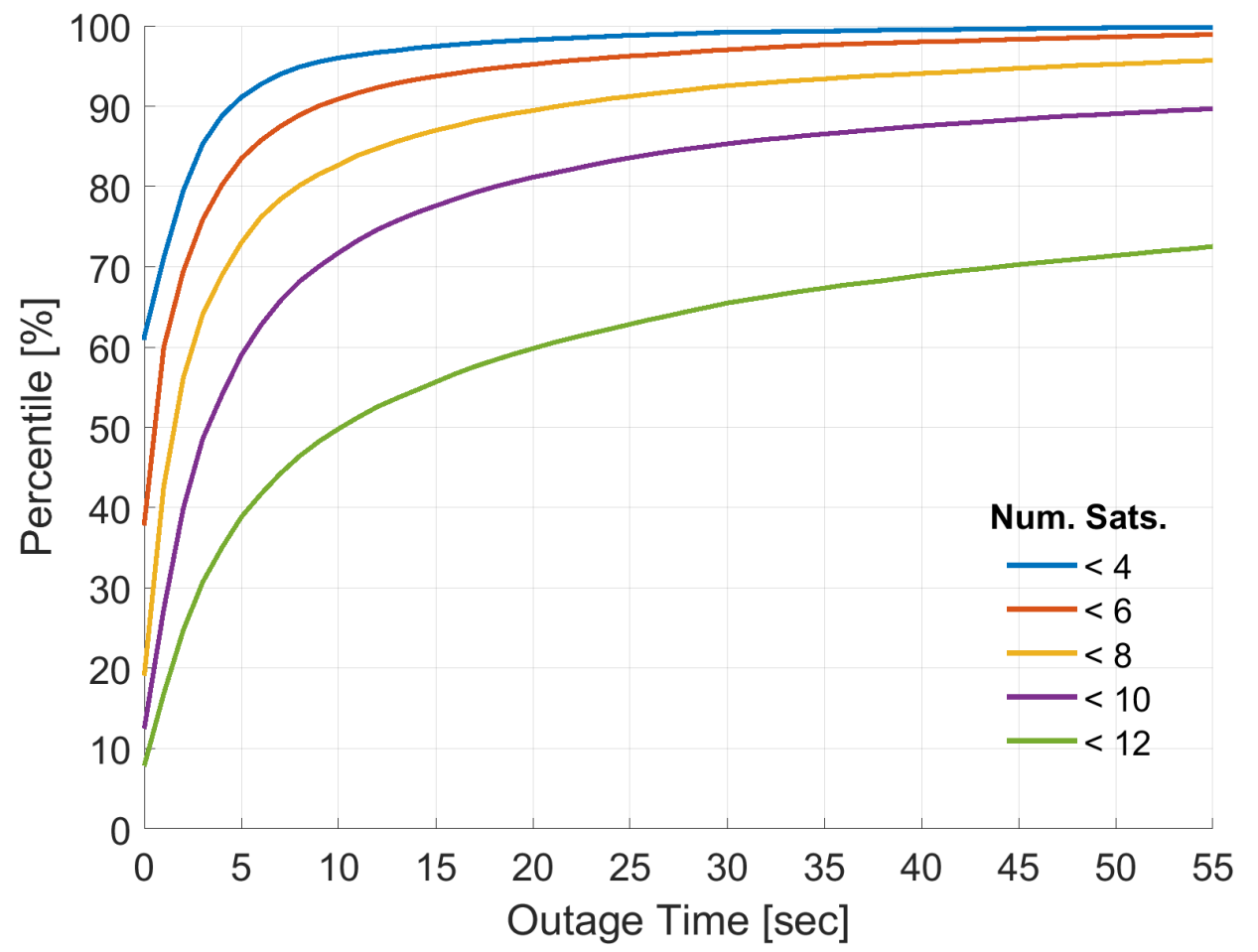

Figure 19: Satellite visibility outage times for the RT3000.

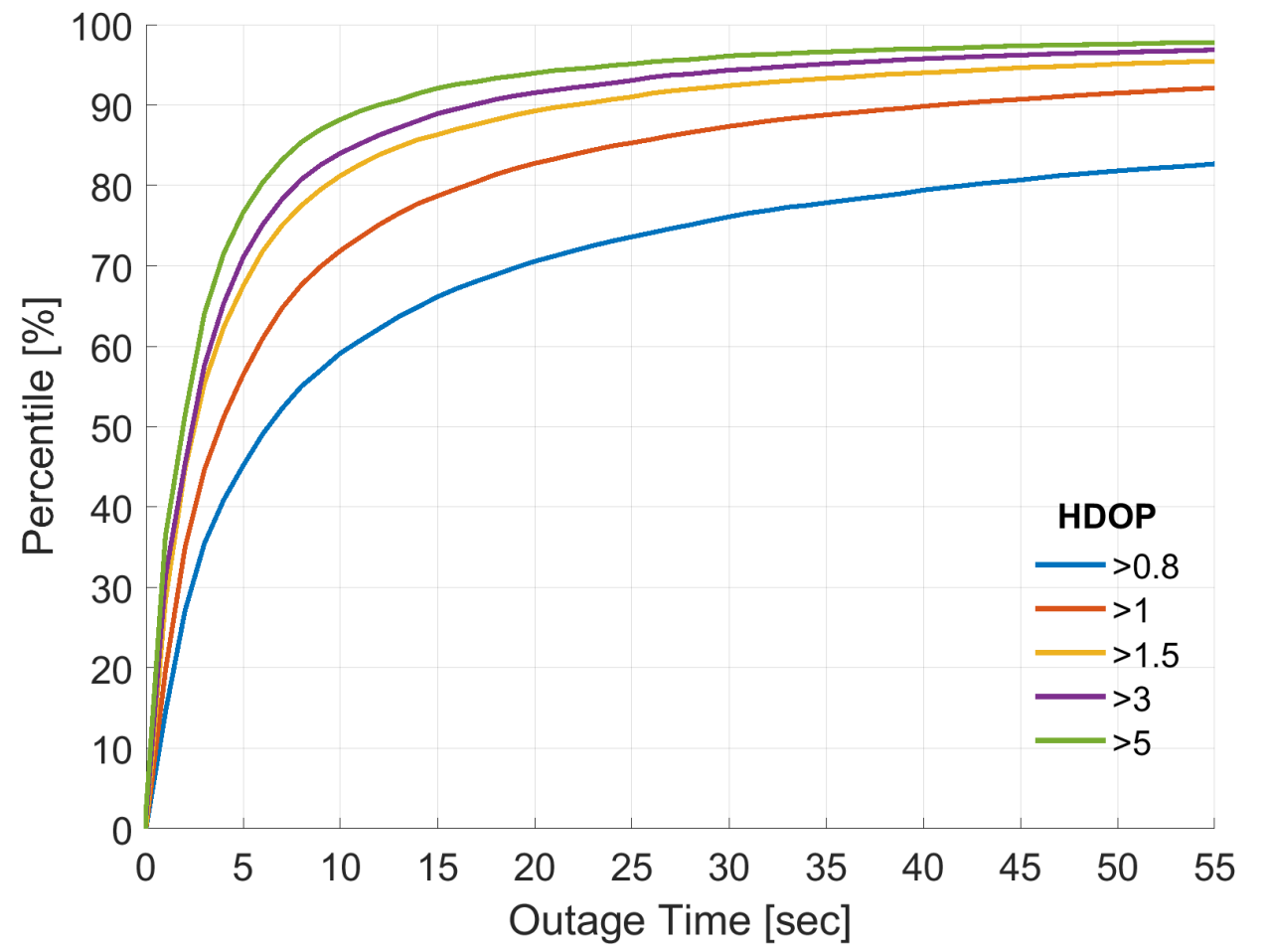

Figure 20: Satellite geometry (HDOP) outage times for the RT3000.

### *RTK Performance*

In this section, the performance of the RTK position solution is assessed. Table 7 shows a summary of position mode and networked RTK correction availability. Position mode availability is a combined measure of environmental factors which cause obstructions to satellites and availability of RTK corrections delivered in real-time over the cellular network. Carrier phase integer ambiguity resolution (RTK fixed) is achieved in nearly 50% of positions. These are the centimeter level positions in the distribution shown in Figure 6. Carrier phase floating point solutions (RTK float) are achieved in 14% of cases. Majority of other data points were reported as standard code phase positioning (33%) with very few being differential code phase (<1%). The latter likely converged quickly to a floating-point carrier phase solution. In less than 3% of cases was no position available where the RT3000 was running in a dead-reckoning mode. Position mode availability is summarized in Figure 21.

Table 7: Summary of position mode and correction availability.

| Position Mode | | | | | Age of Corrections | | |
|---|---|---|---|---|---|---|---|
| RTK Fixed | RTK Float | Diff. Code Phase | Standard Positioning | None | < 2 sec | < 10 sec | < 120 sec |
| 49.9% | 14.1% | 0.3% | 33.0% | 2.7% | 96.0% | 97.3% | 98.3% |

The geospatial distribution of position mode is given in Figure 22. Notice that position mode is not continuous but instead constantly dropping from RTK fixed to SPS or even to no position fix. This is a result of the continuity of satellite visibility and geometry described in the previous section along with availability of RTK corrections.

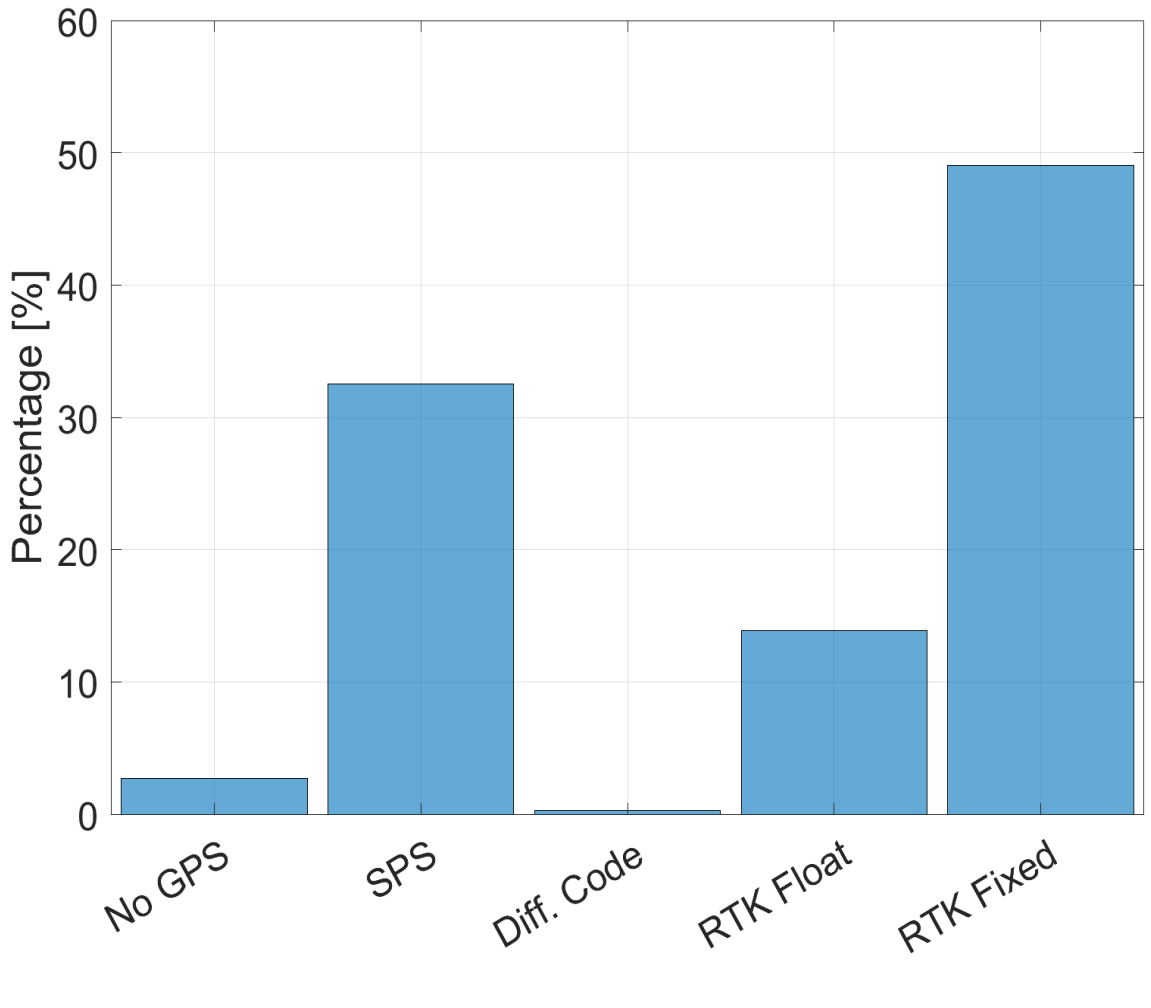

**Figure 21: Availability of RT3000 position modes.**

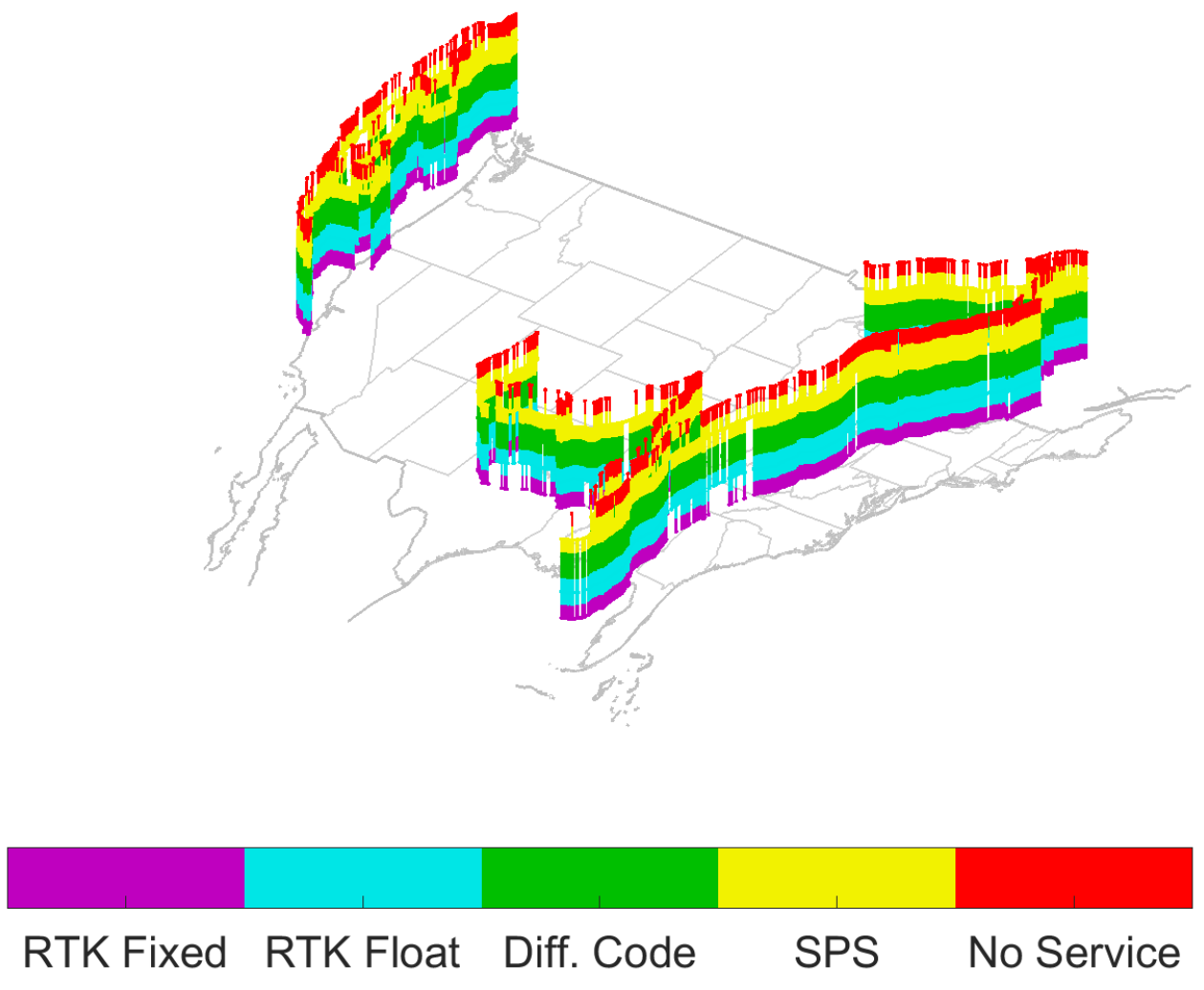

**Figure 22: Geospatial distribution of the RT3000 reported position modes.**

The age of RTK corrections being used by the receiver is summarized in Table 7, where stale corrections would only be used if new ones were unavailable. 'Stale' in the context of networked RTK is typically taken as older than 10 to 15 seconds due to the temporal decorrelation of GNSS satellite, clock, and atmospheric errors between the base station and rover receiver (vehicle) [19], [32]. Since corrections were delivered in real-time, availability is predominantly a measure of cellular coverage along the route. Results show that nearly 96% of corrections are less than 2 seconds old, which is considered to be low latency and hence good performance [19], [32]. In 97% of cases, new corrections could take up to 10 seconds to arrive. In the remaining 3% of cases, corrections could be several minutes old due to extended cellular network outages and be of limited value. The cumulative distribution of correction age is given in Figure 23. The geospatial distribution of correction age is given in Figure 24. This shows that spikes occur mostly in rural areas with poor cellular coverage, e.g. the California State Route 1 along the Pacific Coast. Furthermore, the higher frequency of outages on the west coast is likely due to terrain affecting cellular coverage.

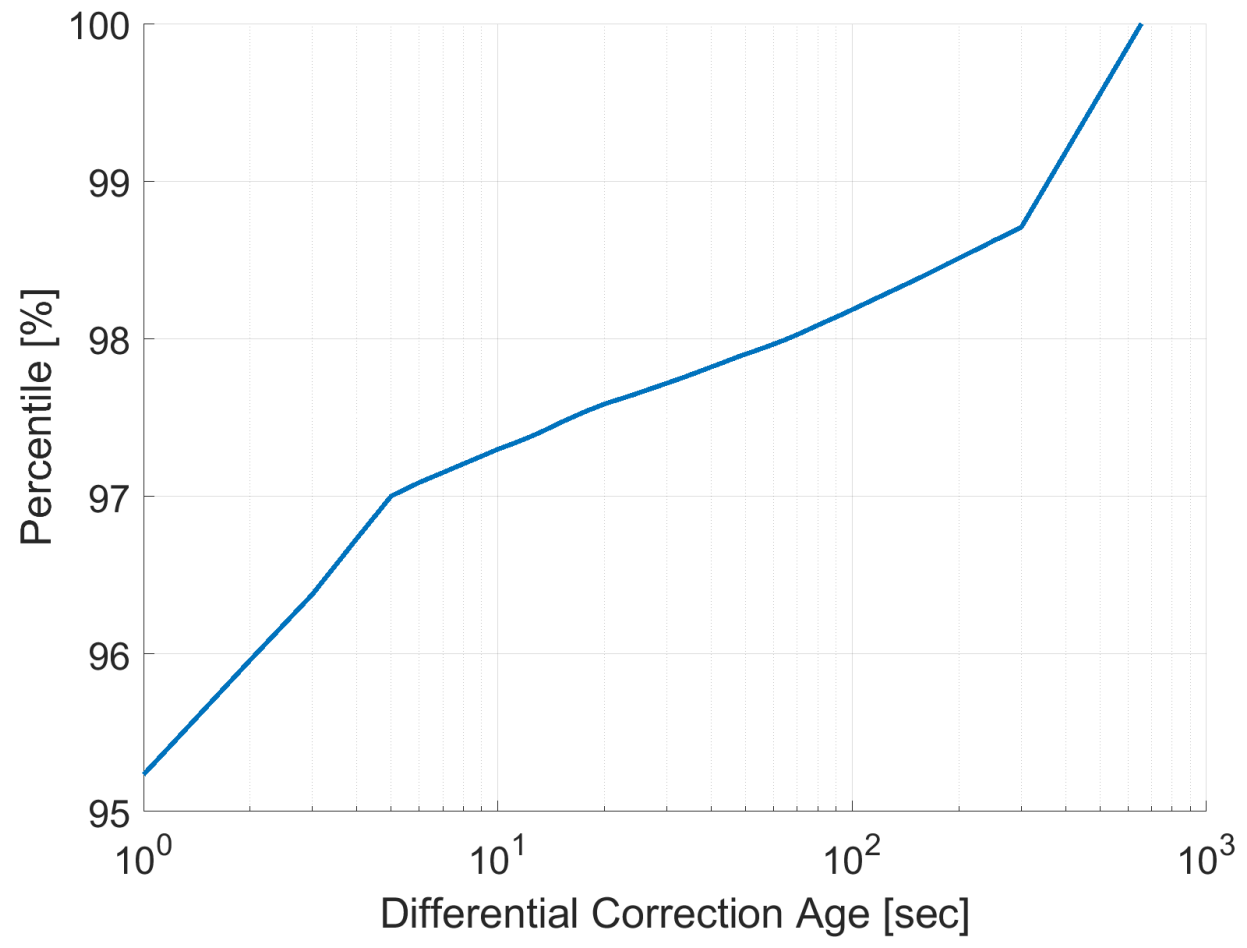

**Figure 23: Empirical cumulative distribution of on-road age of RTK corrections. This is a combined measure of cellular connectivity and RTK correction availability along the route.**

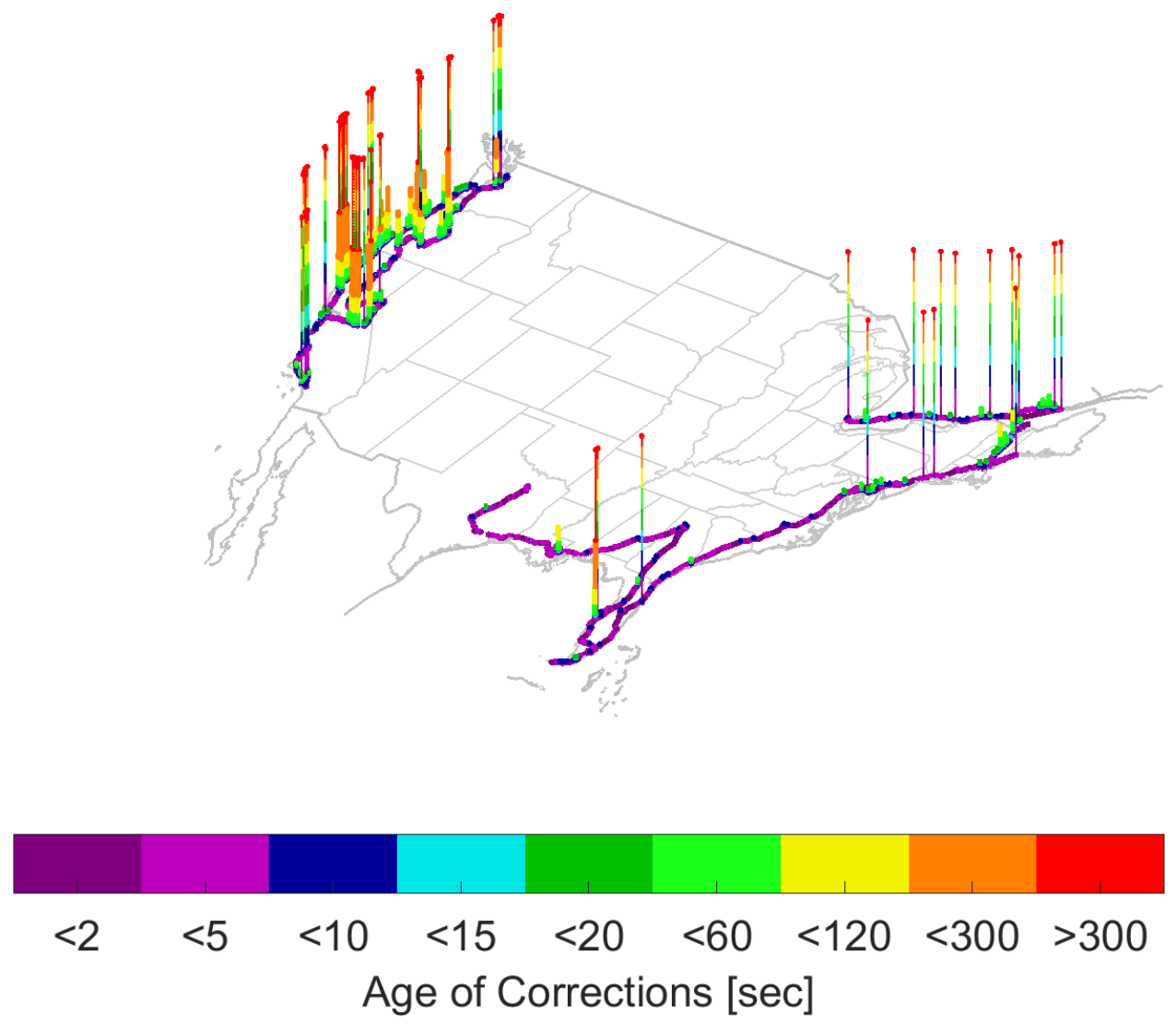

**Figure 24: Geospatial distribution of reported age of RTK corrections. This is a combined measure of cellular connectivity and RTK correction availability along the route.**

The probability of continuity loss of RTK and other position modes is summarized in Table 8 and Figure 25. This shows continuity to be on the same order as that for satellite visibility and geometry as discussed in the previous section. The ranking of solution fragility is shown to be inversely related to accuracy following the order of RTK fixed, RTK float, differential code phase, and standard code phase positioning (SPS) with RTK fixed being most accurate and most transient. This is expected due to the increased difficulty of tracking the carrier phase compared to the code phase used in standard (SPS) positioning. Like satellite visibility and dilution of precision, these again are on the order of $10^{-1}$, a stark difference from the $10^{-6}$ numbers assumed in critical aviation maneuvers [27]. This difference shows one of the challenges facing the development of high integrity GNSS positioning systems on the road.

Figure 26 shows the typical outage times. Median RTK fixed outages are 11 seconds compared to float solutions at 2 seconds. However, in the 95% of cases, outages for RTK fixed and float can be longer than a minute, compared to standard code phase positioning at <7 seconds. Hence, RTK outages can be much longer than the timescales of the critical maneuvers considered here.

**Table 8: Summary of position mode probability of continuity loss.**

| Interval [sec] | RTK Fixed | RTK Float | Differential Code | Standard Positioning |
|---|---|---|---|---|
| 4 | $5.4\times10^{-1}$ | $4.4\times10^{-1}$ | $4.4\times10^{-1}$ | $4.5\times10^{-2}$ |
| 7 | $5.7\times10^{-1}$ | $4.9\times10^{-1}$ | $4.9\times10^{-1}$ | $6.2\times10^{-2}$ |
| 15 | $6.4\times10^{-1}$ | $5.8\times10^{-1}$ | $5.8\times10^{-1}$ | $1.0\times10^{-1}$ |
| 30 | $7.3\times10^{-1}$ | $6.9\times10^{-1}$ | $6.9\times10^{-1}$ | $1.6\times10^{-1}$ |

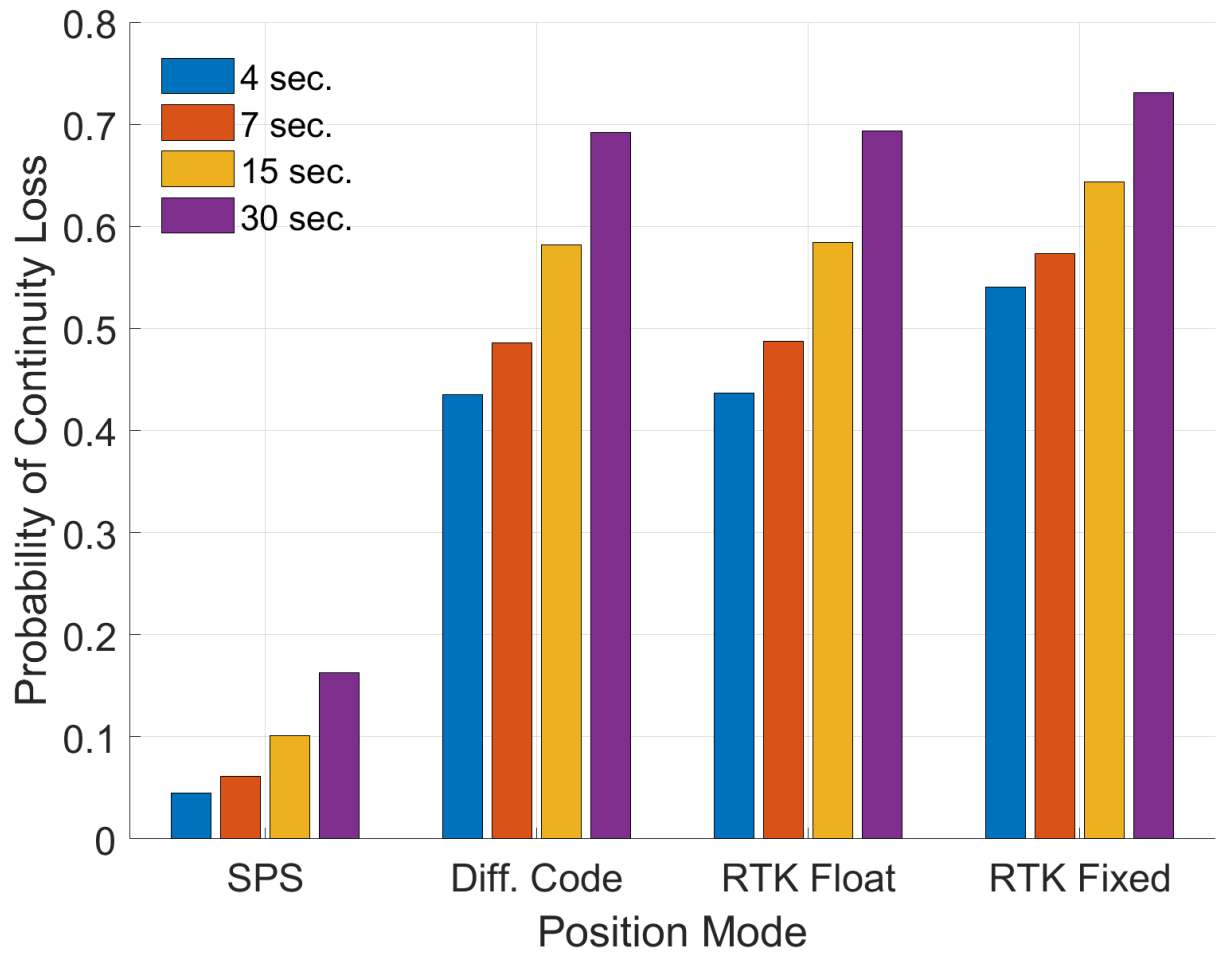

**Figure 25: Position mode probability of continuity loss for the RT3000.**

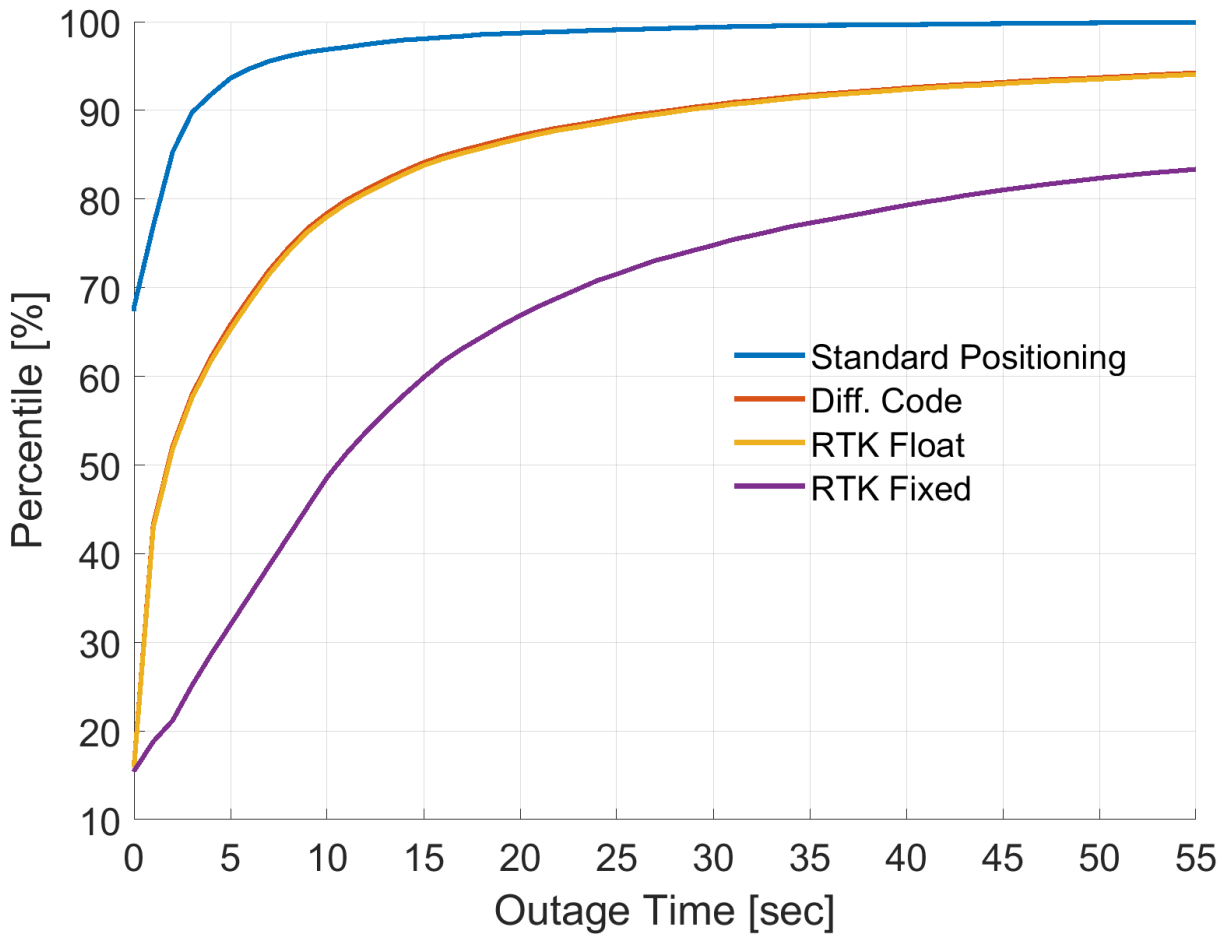

**Figure 26: Position mode outage times for the RT3000.**

***Case Study: San Francisco Bay Area***

To show these factors combined in defining GNSS performance, consider the example of the San Francisco Bay Area in California. Figure 27 shows the position mode map where it is clear that open highways such as the I-280 and US101 have RTK fixed performance with dropouts corresponding to overpasses. The San Mateo Bridge is in open sky and also has very good performance. By comparison, the Richmond-San Rafael Bridge shows poor performance as this was driven eastbound on the lower deck. Downtown San Francisco shows intermittent performance with many dropouts due to obstructions from tall buildings. Figure 28 shows a strong correlation with these events and satellite dropouts. Figure 29 shows that the RTK correction latency was less of a factor, indicating an adequate cellular data connection with the exception of three data points.

Though the bulk measures of accuracy, availability, and continuity presented here are useful for assessing the general performance of GNSS on the road, the geospatial diversity of performance should perhaps not be ignored. Certain road segments could be characterized as having good or bad GNSS performance based on an a-priori map. Availability, continuity, and connectivity could be a layer or feature associated with High Definition (HD) map segments which define the expected performance of GNSS. Such data could be created based on simulation using 3D models of the environment available from HD maps, through fleet vehicle crowd-sourcing, through a calibration or validation driving campaign (seen as an extension of other data-driven automotive calibration processes [33]), or some combination. This can also be built as a byproduct of the HD map making procedure itself since GNSS will be a component of that process assuming mobile mapping vehicles are involved. This GNSS performance map is analogous to the localization layers added to maps for techniques based on relative sensors such as LiDAR [34]. An illustration of what such GNSS performance map layers could look like is given in Figure 30. A GNSS performance map can also inform driving policy. For example, certain maneuvers may want to be limited in certain areas due to poor GNSS performance in order to maintain integrity of the overall virtual driver system. Furthermore, it could inform following distances or lane changes to limit satellite occlusions caused by large trucks or known static buildings in order to operate in friendlier GNSS environments.

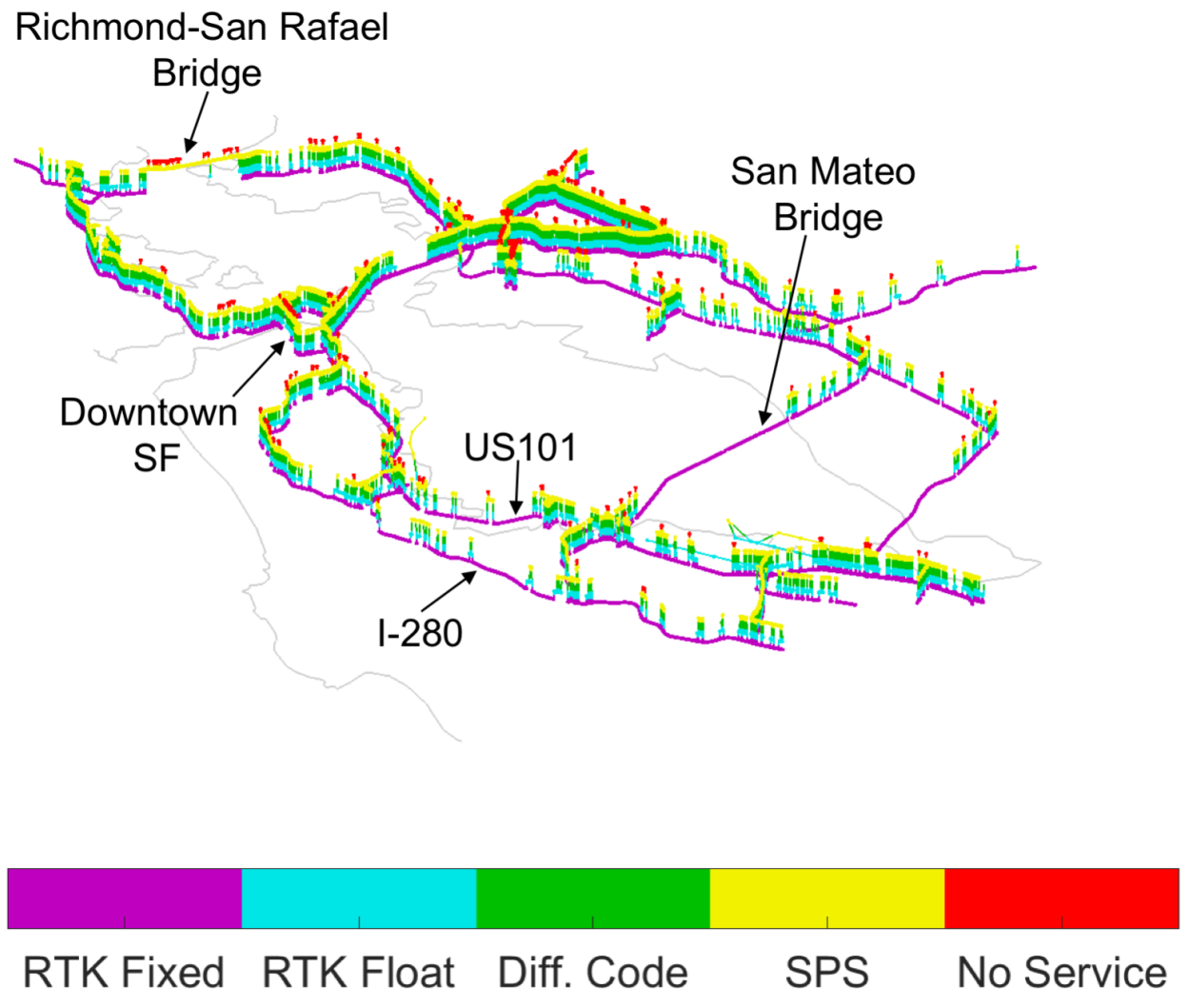

**Figure 27: Geographic distribution of the RT3000 reported position modes in the San Francisco Bay Area.**

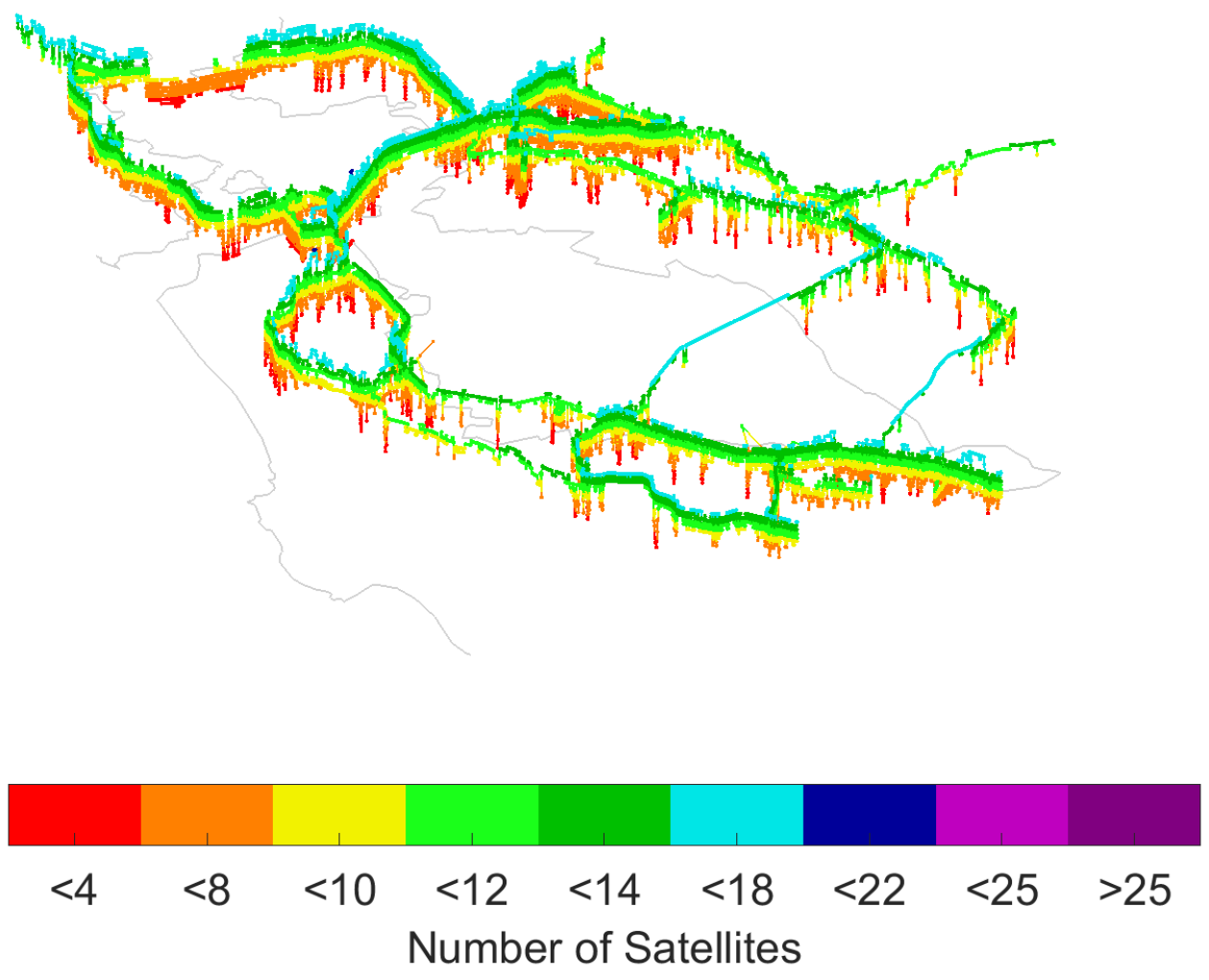

**Figure 28: Geographic distribution of the RT3000 reported satellite visibility in the San Francisco Bay Area.**

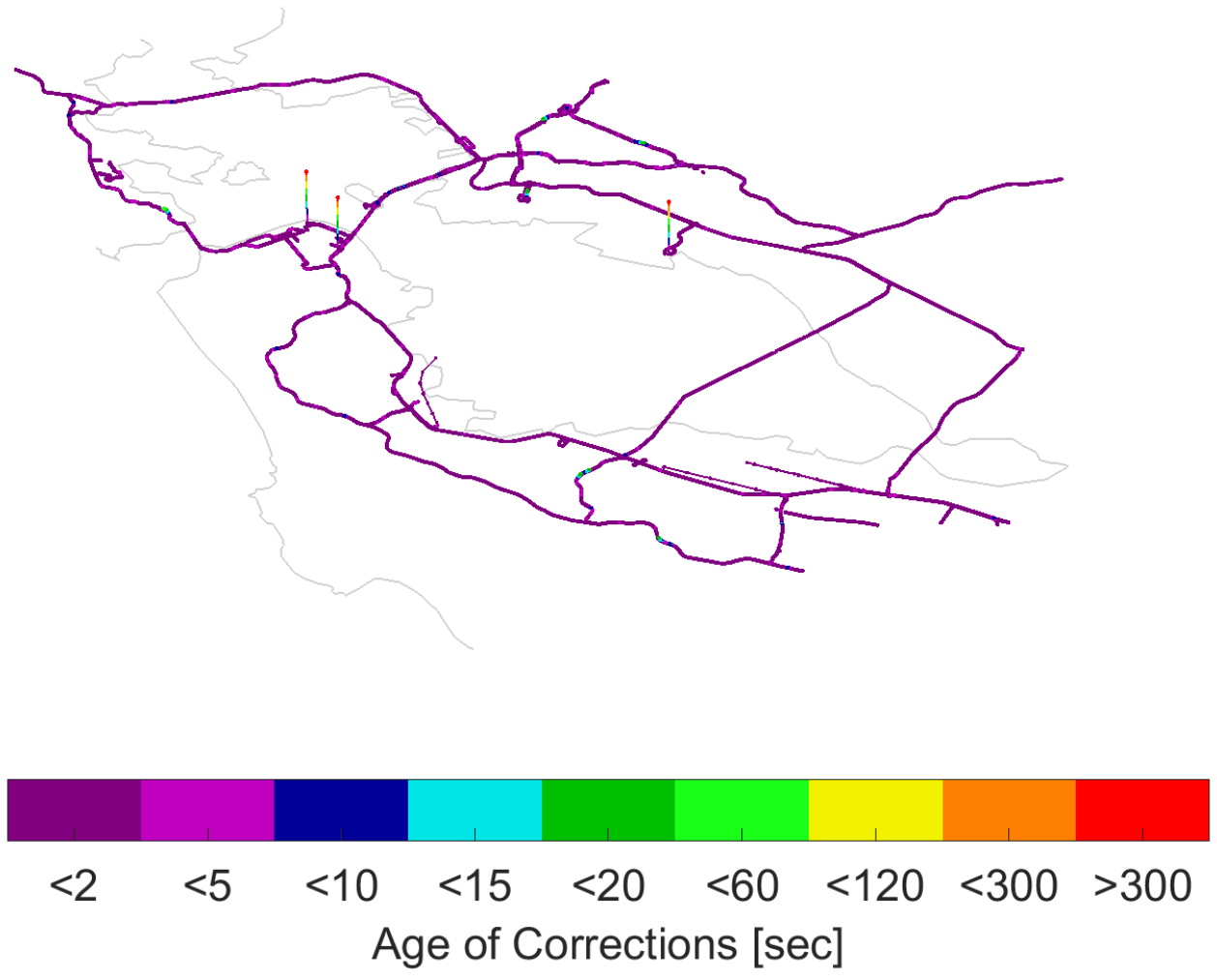

**Figure 29: Geographic distribution of the RT3000 reported age of RTK corrections in the San Francisco Bay Area.**

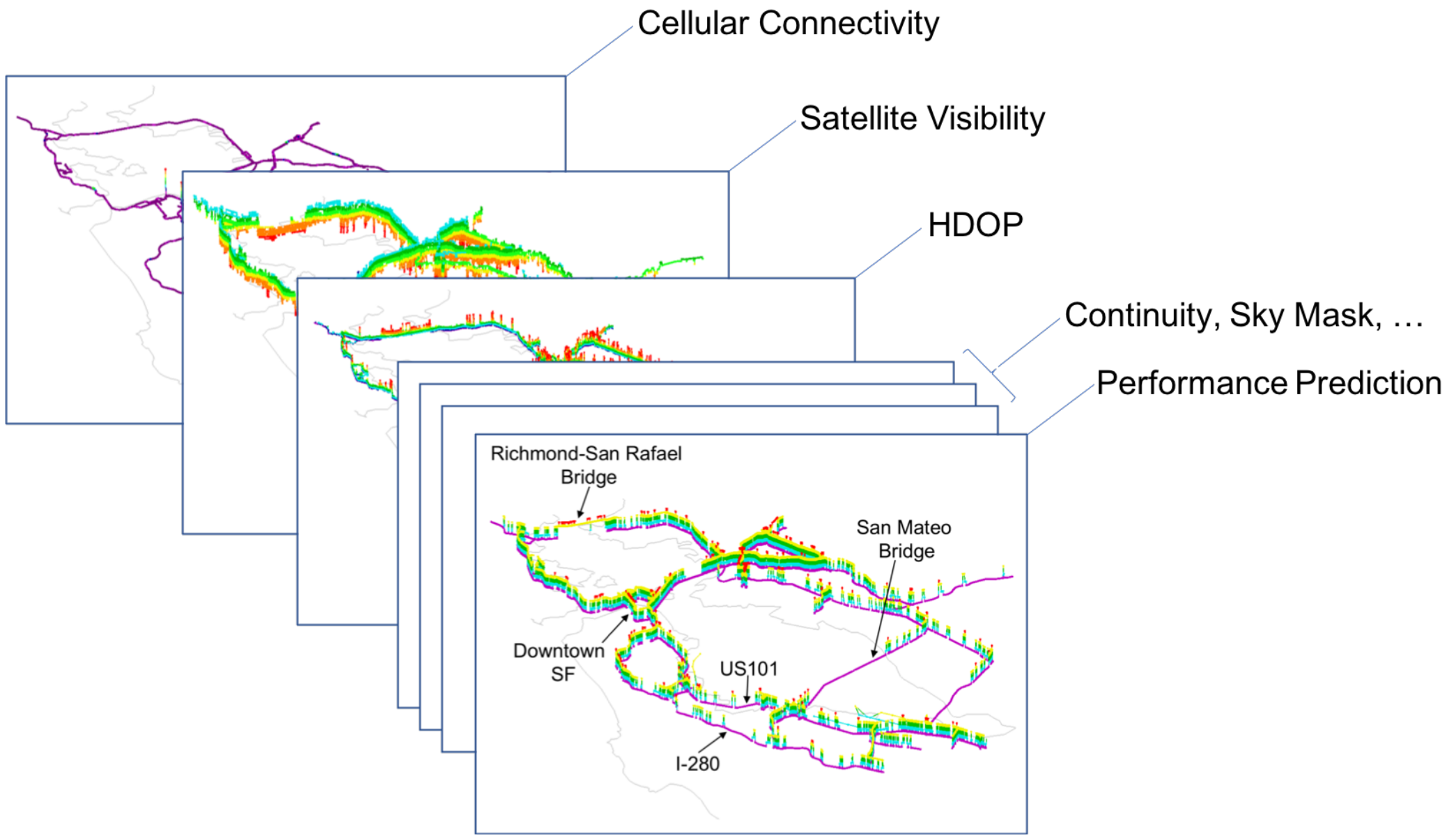

**Figure 30: Example of layers in a GNSS performance map.**

## CONCLUSION

This paper examines the state of GNSS on predominantly highway driving in North America today with stand-alone, single frequency automotive grade receivers and what might be possible in the not-too-distant future with widespread, networked GNSS correction services and multi-frequency. This found the role of current L1-only automotive GNSS best suited to positioning levels required for road determination at 98% <5 meters but less suitable for lane determination at 57%. Multi-frequency receivers with RTK corrections were found more capable with road determination at 99.5%, lane determination at 98%, and highway-level lane departure protection at 91%. RTK correction services are already 97% available on both the east and west coasts of the United States where deficiencies stemmed only from a lack of cellular connectivity, though in general there are still geographic regions that still have an RTK coverage deficiency.

Though open sky accuracy was usually sufficient, availability and continuity proved more difficult due to the regularity of satellite occlusions on the road. With a GPS + GLONASS receiver tracking L1 + L2, RTK fixed positioning was available 50% of the time with a probability of continuity loss of 0.54 over 4 seconds compared to standard code phase positioning at 98% availability with a continuity loss probability of 0.045. Hence, as expected, the RTK solution was more fragile than code phase positioning (SPS). This is also reflected in outage times where in 95% of cases, SPS positioning outages were <7 seconds compared to RTK fixed / float at which could be more than 1 minute. Looking forward, availability and continuity could improve as satellite modernization make more multi-frequency satellites available and as more constellations such as Galileo and BeiDou become fully operationally. However, satellite occlusions encountered in the road environment such as overpasses, signs, and buildings will continue to present challenges.

Overall, the automotive GNSS today appears best suited to supporting ‘which road’ applications such as geofencing whereas RTK or PPP correction enabled receivers may be able to support widescale lane determination and potentially lane departure under favorable GNSS conditions. The empirically derived parameters on accuracy, availability, and continuity are the building block upon which models for GNSS reliability and performance on the road can be built. Though nearly 4 constellations are in operation with GPS, GLONASS, and soon BeiDou and Galileo along with correction infrastructure in place, challenges remain with availability and continuity risk which differ from those found in aviation by orders of magnitude. Continuity loss coupled with the distribution of outage times for different levels of positioning is the basis on which to start the sizing of IMUs and other localization sensors to maintain position confidence during GNSS outages. Furthermore, this leads to questions about defining availability and continuity risk parameters on a per road segment basis and perhaps as an additional layer to the vehicle's map. This GNSS performance parameter map could be derived based on data-driven, simulation, or crowdsourcing approaches and can be an evolving and essential piece of information used in predicting the integrity of GNSS positions along given routes and in turn enabling certain ADAS, AD, and Connected Vehicle features.

**ACKNOWLEDGEMENTS**

The authors would like to greatly thank Ford Motor Company for supporting this work.

**REFERENCES**

1. I. Skog and P. Handel, "In-Car Positioning and Navigation Technologies—A Survey," *IEEE Trans. Intell. Transp. Syst.*, vol. 10, no. 1, pp. 4–21, Mar. 2009.
2. American Association of State Highway and Transportation Officials, "A Policy on Geometric Design of Highways and Streets." 2001.
3. A. Davies, "Thank Maps for the Cadillac CT6's Super Cruise Self-Driving | WIRED," *Wired*, 2018. [Online]. Available: https://www.wired.com/story/cadillac-super-cruise-ct6-lidar-laser-maps/. [Accessed: 04-Jan-2019].
4. C. Hay, "Use of Precise Point Positioning for Cadillac Super Cruise," in *Munich Satellite Navigation Summit*, 2018.
5. C. Hay, "Autonomous Vehicle Localization by Leveraging Cellular Connectivity," in *Symposium on the Future Networked Car 2018 (FNC-2018)*, 2018.
6. S. Stephenson, "Automotive applications of high precision GNSS," University of Nottingham, 2016.
7. C. Basnayake, T. Williams, P. Alves, and G. Lachapelle, *Can GNSS drive V2X?*, vol. 21. 2010.
8. National Highway Traffic Safety Administration and Department of Transportation, "Federal Motor Vehicle Safety Standards; V2V Communications, Docket No. NHTSA–2016–0126, RIN 2127–AL55," Federal Register, Washington, DC, 2017.
9. T. G. R. Reid, S. E. Houts, R. Cammarata, G. Mills, S. Agarwal, A. Vora, and G. Pandey, "Localization Requirements for Autonomous Vehicles," *SAE Int. J. Connect. Auton. Veh.*, vol. Submitted, 2019.
10. T. G. R. Reid, S. E. Houts, R. Cammarata, G. Mills, S. Agarwal, A. Vora, and G. Pandey, "Localization Requirements for Autonomous Vehicles," *arXiv:1906.01061*, Jun. 2019.
11. National Highway Traffic Safety Administration, "Vehicle Safety Communications-Applications (VSC-A) Final Report: Appendix Volume 2 Communications and Positioning (DOT HS 811 492C)," Washington, DC, 2011.
12. T. Williams, P. Alves, G. Lachapelle, and C. Basnayake, "Evaluation of GPS-based methods of relative positioning for automotive safety applications," *Transp. Res. Part C Emerg. Technol.*,

vol. 23, pp. 98–108, Aug. 2012.
13. T. Pilutti and C. Wallis, "Onroad GPS Availability Analysis," in *SAE 2010 World Congress & Exhibition*, 2010.
14. S. Thrun, M. Montemerlo, H. Dahlkamp, D. Stavens, A. Aron, J. Diebel, P. Fong, J. Gale, M. Halpenny, G. Hoffmann, K. Lau, C. Oakley, M. Palatucci, V. Pratt, P. Stang, S. Strohband, C. Dupont, L.-E. Jendrossek, C. Koelen, C. Markey, C. Rummel, J. van Niekerk, E. Jensen, P. Alessandrini, G. Bradski, B. Davies, S. Ettinger, A. Kaehler, A. Nefian, and P. Mahoney, "Stanley: The robot that won the DARPA Grand Challenge," *J. F. Robot.*, vol. 23, no. 9, pp. 661–692, Sep. 2006.
15. L. de Groot, E. Infante, A. Jokinen, B. Kruger, and L. Norman, "Precise Positioning for Automotive with Mass Market GNSS Chipsets," in *Proceedings of the 31st International Technical Meeting of The Satellite Division of the Institute of Navigation (ION GNSS+ 2018)*, 2018, pp. 596–610.
16. U. Niesen, J. Jose, and X. Wu, "Accurate Positioning in GNSS-Challenged Environments with Consumer-Grade Sensors," in *Proceedings of the 31st International Technical Meeting of The Satellite Division of the Institute of Navigation (ION GNSS+ 2018)*, 2018, pp. 503–514.
17. S. Shade and P. H. Madhani, "Android GNSS Measurements - Inside the BCM47755," in *Proceedings of the 31st International Technical Meeting of The Satellite Division of the Institute of Navigation (ION GNSS+ 2018)*, 2018, pp. 554–579.
18. Y. Lu, S. Ji, W. Chen, and Z. Wang, "Assessing the Performance of Raw Measurement from Different Types of Smartphones," in *Proceedings of the 31st International Technical Meeting of The Satellite Division of the Institute of Navigation (ION GNSS+ 2018)*, 2018, pp. 304–322.
19. S. Vana, J. Aggrey, S. Bisnath, R. Leandro, L. Urquhart, and P. Gonzalez, "Analysis of GNSS Correction Data Standards for the Automotive Market," in *Proceedings of the 31st International Technical Meeting of The Satellite Division of the Institute of Navigation (ION GNSS+ 2018)*, 2018, pp. 4197–4214.
20. U. Weinbach, M. Brandl, X. Chen, H. Landau, F. Pastor, N. Reussner, and C. Rodriguez-Solano, "Integrity of the Trimble CenterPoint RTX Correction Service," in *Proceedings of the 31st International Technical Meeting of The Satellite Division of the Institute of Navigation (ION GNSS+ 2018)*, 2018, pp. 1902–1909.
21. S. Banville, M. Bavaro, S. Carcanague, A. Cole, K. Dade, P. Grgich, A. Kleeman, and B. Segal, "Network Modelling Considerations for Wide-area Ionospheric Corrections," in *Proceedings of the 31st International Technical Meeting of The Satellite Division of the Institute of Navigation (ION GNSS+ 2018)*, 2018, pp. 1883–1892.
22. A. Jokinen, C. Ellum, I. Webster, S. Shanmugam, and K. Sheridan, "NovAtel CORRECT with Precise Point Positioning (PPP): Recent Developments," in *Proceedings of the 31st International Technical Meeting of The Satellite Division of the Institute of Navigation (ION GNSS+ 2018)*, 2018, pp. 1866–1882.
23. Federal Highway Administration, "Highway Statistics 2017," 2017. [Online]. Available: https://www.fhwa.dot.gov/policyinformation/statistics/2017/. [Accessed: 26-Feb-2019].
24. U.S. Department of Transportation, Federal Highway Administration, and Federal Transit Administration, "2015 Status of the Nation's Highways, Bridges, and Transit: Conditions & Performance," Washington, DC, 2015.
25. U. Klehmet, T. Herpel, K.-S. Hielscher, and R. German, "Delay Bounds for CAN Communication in Automotive Applications," in *14th GI/ITG Conference - Measurement, Modelling and Evalutation of Computer and Communication Systems*, 2008.
26. T. Herpel, K.-S. Hielscher, U. Klehmet, and R. German, "Stochastic and deterministic

performance evaluation of automotive CAN communication," *Comput. Networks*, vol. 53, no. 8, pp. 1171–1185, Jun. 2009.
27. ICAO, "Annex 10 to the Convention on International Civil Aviation, Volume I, Radio Navigation Aids," Montreal, 2006.
28. B. S. Pervan, "Navigation Integrity for Aircraft Precision Landing Using the Global Positioning System," Stanford University, 1996.
29. T. Toledo and D. Zohar, "Modeling Duration of Lane Changes," *Transp. Res. Rec. J. Transp. Res. Board*, vol. 1999, no. 1, pp. 71–78, Jan. 2007.
30. E. I. Vlahogianni, "Modeling duration of overtaking in two lane highways," *Transp. Res. Part F Traffic Psychol. Behav.*, vol. 20, pp. 135–146, Sep. 2013.
31. A. Eriksson and N. A. Stanton, "Takeover Time in Highly Automated Vehicles: Noncritical Transitions to and From Manual Control," *Hum. Factors J. Hum. Factors Ergon. Soc.*, vol. 59, no. 4, pp. 689–705, Jun. 2017.
32. J. Song, B. Park, and C. Kee, "A Study on GPS/GLONASS Compact Network RTK and Analysis on Temporal Variations of Carrier Phase Corrections for Reducing Broadcast Bandwidth," in *Proceedings of the ION 2017 Pacific PNT Meeting*, 2017, pp. 659–669.
33. N. Pervez, A. K. Kue, A. Appukuttan, J. Bogema, and M. Van Nieuwstadt, "Data Driven Calibration Approach," *SAE Int. J. Commer. Veh.*, vol. 10, no. 1, pp. 353–359, Mar. 2017.
34. J. Levinson, M. Montemerlo, and S. Thrun, "Map-Based Precision Vehicle Localization in Urban Environments."

## APPENDIX

In this appendix, the process of solving for global and body-fixed frame offsets and rotations between two separate localization systems is given. On the vehicle, it is common that the point of localization differs between ground truth (e.g. RT3000) and the localizer under evaluation, in this case a production-grade automotive GNSS. This can be visualized in Figure 3 where the production-grade GNSS has a different antenna phase center than the RT3000 which additionally calculates its position with respect to the center of its IMU. This offset must be accounted for or it will result in errors in the global frame. This can be measured directly with precise metrology equipment as part of calibration, but this section will show how to solve for these offsets based on the data.

In addition to body-fixed offsets, it is possible that the local Cartesian frame of the ground truth reference requires alignment with that of the localizer under evaluation. This transformation is at most affine, meaning that it could involve a translation and rotation to align the frames. Since both the RT3000 and production-grade GNSS were operating in WGS84, this offset should be negligible, but it will be included in the derivation for completeness.

Each ground truth reference position $\mathbf{x}_{\mathrm{ref},i}$ collected at time $t_i$ can be written as a function of the corresponding position of the localizer under evaluation $\mathbf{x}_{\mathrm{eval},i}$, in this case the production-grade GNSS, as follows:

$$\mathbf{x}_{\mathrm{ref},i} = \mathbf{R}_{\mathrm{eval}}^{\mathrm{ref}}\,\mathbf{x}_{\mathrm{eval},i} + \mathbf{R}_{\mathrm{body},i}^{\mathrm{ref}}\,\mathbf{y}_{\mathrm{body}} + \mathbf{y}_{\mathrm{eval}} + \boldsymbol{\epsilon}_i \tag{1}$$

where $\mathbf{R}_{\mathrm{eval}}^{\mathrm{ref}}$ is the unknown fixed rotation matrix transforming the frame of the localizer to the ground truth frame, $\mathbf{y}_{\mathrm{eval}}$ is the unknown fixed offset between the global ground truth and global localizer frames, $\mathbf{y}_{\text{body}}$ is the unknown body-fixed offset between the center of positioning of the localizer and ground truth expressed in the body frame, $\mathbf{R}_{\text{body},i}^{\text{ref}}$ is the known rotation matrix at time $t_i$ transforming body-fixed coordinates to ground truth global coordinates as measured by the ground truth reference, and $\boldsymbol{\epsilon}_i$ is the remaining error at time $t_i$ to be characterized.

We wish to set up the problem so that we can solve for $\mathbf{R}_{\text{eval}}^{\text{ref}}$, $\mathbf{y}_{\text{body}}$, and $\mathbf{y}_{\text{eval}}$. We will begin by expanding the items that are known. The known transformation between the body and ground truth frame $\mathbf{R}_{\text{body},i}^{\text{ref}}$ is as follows:

$$\mathbf{R}_{\text{body},i}^{\text{ref}} = \mathbf{R}_3(\psi_i)\mathbf{R}_2(\theta_i)\mathbf{R}_1(\phi_i) \tag{2}$$

This is known as a 3-2-1 Euler sequence where $\psi_i$, $\theta_i$, and $\phi_i$ are the yaw (heading), pitch, and roll angles at time $t_i$, respectively. These define the orientation of the vehicle with respect to the North-East-Down (NED) coordinate frame, and are an output of the ground truth system. The OxTS RT3000 system uses this NED definition of the angles though there are others. For reference, these elemental rotation matrices are defined as follows:

$$\mathbf{R}_1(\phi_i) = \begin{bmatrix} 1 & 0 & 0 \\ 0 & \cos\phi_i & -\sin\phi_i \\ 0 & \sin\phi_i & \cos\phi_i \end{bmatrix} \tag{3}$$

$$\mathbf{R}_2(\theta_i) = \begin{bmatrix} \cos\theta_i & 0 & \sin\theta_i \\ 0 & 1 & 0 \\ -\sin\theta_i & 0 & \cos\theta_i \end{bmatrix} \tag{4}$$

$$\mathbf{R}_3(\psi_i) = \begin{bmatrix} \cos\psi_i & -\sin\psi_i & 0 \\ \sin\psi_i & \cos\psi_i & 0 \\ 0 & 0 & 1 \end{bmatrix} \tag{5}$$

The next step is to expand the following matrix multiplication:

$$\mathbf{R}_{\text{eval}}^{\text{ref}}\,\mathbf{x}_{\text{eval},i} = \begin{bmatrix} R_{11} & R_{12} & R_{13} \\ R_{21} & R_{22} & R_{23} \\ R_{31} & R_{32} & R_{33} \end{bmatrix} \begin{bmatrix} x_i \\ y_i \\ z_i \end{bmatrix} = \begin{bmatrix} R_{11}x_i + R_{12}y_i + R_{13}z_i \\ R_{21}x_i + R_{22}y_i + R_{23}z_i \\ R_{31}x_i + R_{32}y_i + R_{33}z_i \end{bmatrix} \tag{6}$$

This can be equivalently written as the following product:

$$\mathbf{R}_{\text{eval}}^{\text{ref}}\,\mathbf{x}_{\text{eval},i} = \mathbf{X}_{\text{eval},i}\mathbf{r}_{\text{eval}}^{\text{ref}} = \begin{bmatrix} x_i & y_i & z_i & 0 & 0 & 0 & 0 & 0 & 0 \\ 0 & 0 & 0 & x_i & y_i & z_i & 0 & 0 & 0 \\ 0 & 0 & 0 & 0 & 0 & 0 & x_i & y_i & z_i \end{bmatrix} \begin{bmatrix} R_{11} \\ R_{12} \\ R_{13} \\ R_{21} \\ R_{22} \\ R_{23} \\ R_{31} \\ R_{32} \\ R_{33} \end{bmatrix} \tag{7}$$

Neglecting the error term, this allows us to re-write Eqn (1) as:

$$\mathbf{x}_{\text{ref},i} = \mathbf{X}_{\text{eval},i}\mathbf{r}_{\text{eval}}^{\text{ref}} + \mathbf{R}_{\text{body},i}^{\text{ref}}\,\mathbf{y}_{\text{body}} + \mathbf{y}_{\text{eval}} \tag{8}$$

$$\mathbf{x}_{\text{ref},i} = \begin{bmatrix} \mathbf{X}_{\text{eval},i} & \mathbf{R}_{\text{body},i}^{\text{ref}} & \mathbf{I}_{3\times3} \end{bmatrix} \begin{bmatrix} \mathbf{r}_{\text{eval}}^{\text{ref}} \\ \mathbf{y}_{\text{body}} \\ \mathbf{y}_{\text{eval}} \end{bmatrix} \tag{9}$$

where $\mathbf{I}_{3\times3}$ is the 3×3 identity matrix. This relationship holds for all collected data points, allowing us to stack them as follows:

$$\begin{bmatrix} \mathbf{x}_{\text{ref},1} \\ \mathbf{x}_{\text{ref},2} \\ \vdots \\ \mathbf{x}_{\text{ref},N} \end{bmatrix} = \begin{bmatrix} \mathbf{X}_{\text{eval},1} & \mathbf{R}_{\text{body},1}^{\text{ref}} & \mathbf{I}_{3\times3} \\ \mathbf{X}_{\text{eval},2} & \mathbf{R}_{\text{body},2}^{\text{ref}} & \mathbf{I}_{3\times3} \\ \vdots & \vdots & \vdots \\ \mathbf{X}_{\text{eval},N} & \mathbf{R}_{\text{body,N}}^{\text{ref}} & \mathbf{I}_{3\times3} \end{bmatrix} \begin{bmatrix} \mathbf{r}_{\text{eval}}^{\text{ref}} \\ \mathbf{y}_{\text{body}} \\ \mathbf{y}_{\text{eval}} \end{bmatrix} \tag{10}$$

This now has the form $\mathbf{b} = \mathbf{Az}$ where we desire to find the best estimate $\hat{\mathbf{z}}$. One way to do this is to find the $\mathbf{z}$ that minimizes the Root-Mean-Squared (RMS) error $||\mathbf{Az} - \mathbf{b}||_2$. The solution to the least-squares problem is the Moore Penrose pseudo-inverse given by:

$$\hat{\mathbf{z}} = \mathbf{A}^{\dagger}\mathbf{b} = (\mathbf{A}^T\mathbf{A})^{-1}\mathbf{A}^T\mathbf{b} \tag{11}$$

With estimates of the rotation and offset parameters $\hat{\mathbf{r}}_{\text{eval}}^{\text{ref}}$, $\hat{\mathbf{y}}_{\text{body}}$, and $\hat{\mathbf{y}}_{\text{eval}}$, the residuals are the errors $\boldsymbol{\epsilon}_i$ to be characterized:

$$\begin{bmatrix} \boldsymbol{\epsilon}_1 \\ \boldsymbol{\epsilon}_2 \\ \vdots \\ \boldsymbol{\epsilon}_N \end{bmatrix} = \begin{bmatrix} \mathbf{x}_{\text{ref},1} \\ \mathbf{x}_{\text{ref},2} \\ \vdots \\ \mathbf{x}_{\text{ref},N} \end{bmatrix} - \begin{bmatrix} \mathbf{X}_{\text{eval},1} & \mathbf{R}_{\text{body},1}^{\text{ref}} & \mathbf{I}_{3\times3} \\ \mathbf{X}_{\text{eval},2} & \mathbf{R}_{\text{body},2}^{\text{ref}} & \mathbf{I}_{3\times3} \\ \vdots & \vdots & \vdots \\ \mathbf{X}_{\text{eval},N} & \mathbf{R}_{\text{body,N}}^{\text{ref}} & \mathbf{I}_{3\times3} \end{bmatrix} \begin{bmatrix} \hat{\mathbf{r}}_{\text{eval}}^{\text{ref}} \\ \hat{\mathbf{y}}_{\text{body}} \\ \hat{\mathbf{y}}_{\text{eval}} \end{bmatrix} \tag{12}$$

In this form, these errors are expressed in the coordinates of the ground truth system, which in the case of the OxTS RT is NED. Though global errors are a useful characterization, it is also helpful to project errors into the vehicle's lateral and longitudinal directions. Lateral errors tell us how close we are getting to the lane lines and longitudinal errors how well we know our position along the lane. These errors are determined by projecting $\boldsymbol{\epsilon}_i$ into the vehicle body frame as follows:

$$\begin{bmatrix} \epsilon_{\text{longitdinal},i} \\ \epsilon_{\text{lateral},i} \\ \epsilon_{\text{vertical},i} \end{bmatrix} = \mathbf{R}_{\text{body},i}^{\text{ref}}\,\boldsymbol{\epsilon}_i = \mathbf{R}_{\text{ref},i}^{\text{body}^T}\,\boldsymbol{\epsilon}_i \tag{13}$$